\documentclass[]{fairmeta}

\usepackage{amsmath}
\usepackage{amssymb}
\usepackage{booktabs}
\usepackage{multirow}
\usepackage{array}
\usepackage{tabularx}
\usepackage{pifont}
\usepackage{xspace}
\usepackage{adjustbox}
\usepackage{makecell}
\usepackage{enumitem}
\usepackage{dblfloatfix}
\newcommand{\Bench}{\textsc{EgoGenEval}\xspace}
\newcommand{\TrainSet}{\textsc{EgoGen-Train}\xspace}

\newcommand{\cmark}{\ding{51}}
\newcommand{\xmark}{\ding{55}}
\newcommand{\pmark}{$\triangle$}

\title{Beyond Visual Quality: Evaluating Physical Consistency under Ego-Motion with \Bench{}}

\author[1,2]{Yilin Long}
\author[1,3]{Chenming Zhu}
\author[1,2]{Zitang Gou}
\author[1,4]{Jingli Lin}
\author[1,\ddagger]{Tai Wang}
\affiliation[1]{Shanghai AI Laboratory}
\affiliation[2]{Fudan University}
\affiliation[3]{The University of Hong Kong}
\affiliation[4]{Shanghai Jiao Tong University}
\contribution[\ddagger]{Corresponding author}

\hypersetup{
  pdftitle={Beyond Visual Quality: Evaluating Physical Consistency under Ego-Motion with EgoGenEval},
  pdfauthor={Yilin Long, Chenming Zhu, Zitang Gou, Jingli Lin, Tai Wang},
  pdfsubject={Benchmarking camera-motion grounding and scene-state preservation in pose-free visual generation},
  pdfkeywords={visual generation, ego-motion, physical consistency, benchmark, world models}
}

\metadata[Code]{\url{https://github.com/InternRobotics/EgoGenEval}}

\abstract{Recent visual generators produce high-fidelity images yet often violate physical consistency under ego-motion, limiting their use for spatial reasoning and embodied planning. Existing benchmarks largely focus on isolated images or single-step quality, leaving this challenge underexplored. We introduce \Bench{}, a geometry-grounded, pose-free benchmark designed to evaluate the physical consistency of visual generators under ego-motion, and organize our study into two parts. (1) \Bench{} contains 1,400 cases and 2,360 target views spanning single-step and multi-step ego-motion. It separately measures Camera Motion Grounding (CMG) and Scene State Preservation (SSP), with both metrics validated against blinded human judgments. Evaluating 16 pose-free generators together with two pose-conditioned references reveals that current models struggle to execute camera motion while maintaining scene state, and that no system performs well on both axes at once. (2) To examine whether benchmark-derived data can improve these capabilities, we build \TrainSet{} from the same geometry-grounded pipeline and run controlled SFT studies. These show that pairwise supervision does not reliably improve camera-motion grounding and scene-state preservation together: even at the full training pool and the longest budget, scene preservation gains a fraction of what camera motion does. This points to the pairwise teacher-forced objective itself as the binding constraint, motivating a trajectory-centric paradigm that couples self-conditioned rollouts with explicit pose and visibility supervision.}

\begin{document}

\maketitle

\section{Introduction}
\label{sec:intro}

Recent visual generators offer increasingly photorealistic and controllable synthesis~\citep{saharia2022photorealistic,he2025cameractrl}, motivating their use as \emph{visual simulators} for spatial reasoning and agentic visual imagination~\citep{yang2026mindjourney,zhu2026thinkingimaginationagenticvisual}. For such uses, the generator should execute the instructed camera motion while preserving object presence, spatial layout, and appearance across views. However, a plausible output can drift toward an input view, mis-scale the requested motion, or lose objects during a rollout. We therefore ask: \emph{how reliably do pose-free visual generators ground natural-language camera motions while preserving scene state over short rollouts?}

Existing benchmarks only partially address this question. They emphasize isolated-image composition, single-image editing, explicit pose or trajectory control, or holistic world-model scores~\citep{ghosh2023geneval,huang2023t2i,wang2026placebenchmarkingspatialintelligence,he2025cameractrl,duan2025worldscore,ying2026wbenchcomprehensivemultiturnbenchmark,xiao2026spatialeditbenchmarkingfinegrainedimage}. No existing setting evaluates both capabilities in a unified pose-free rollout protocol while reporting motion realization and target-view scene preservation as separate outcomes. Table~\ref{tab:benchmark_scope} summarizes this distinction.

\begin{figure*}[t!]
\centering
\includegraphics[width=\textwidth]{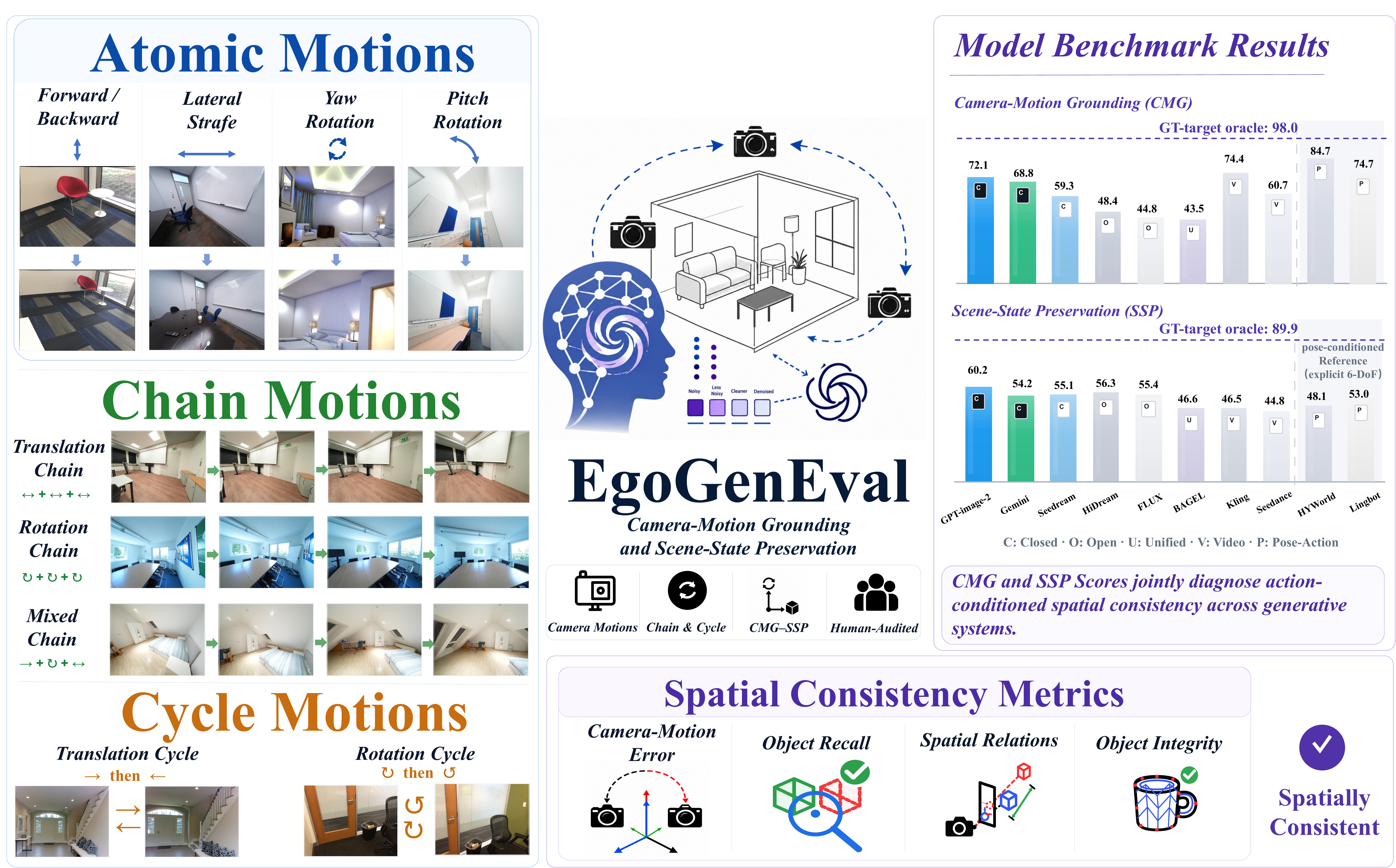}
\caption{Overview of \Bench{}. Atomic, Chain, and inverse-Cycle protocols evaluate magnitude-specified camera motions under controlled visual context (left). CMG and SSP separately measure camera-motion execution and target-view scene-state preservation using geometry- and object-centric components (center and bottom). Representative leaderboard results illustrate the distinct performance profiles induced by the two axes (right).}
\label{fig:teaser}
\end{figure*}

To address this gap, we introduce \Bench{}, comprising 1,400 cases and 2,360 target views from posed RGB-D indoor scenes. It spans four atomic motions, three-step chains, inverse cycles, and $K{=}1$--$4$ input-view settings (\mbox{Figure~\ref{fig:teaser}}). Evaluated models receive only images and magnitude-specified natural-language instructions. We measure \emph{camera-motion fidelity} with the Camera Motion Grounding Score (CMG-Score) and \emph{environment fidelity} with the Scene State Preservation Score (SSP-Score), while reporting conventional image metrics only as auxiliary diagnostics. Both scores closely track aggregate rankings from blinded human evaluations across six representative systems.

Our evaluation of 16 pose-free systems, contextualized by two pose-conditioned references, reveals a consistent \emph{motion--state score gap}. No evaluated system excels at both camera-motion grounding and scene-state preservation, and similar Overall scores can conceal markedly different CMG--SSP profiles. Within camera-motion grounding, directional compliance is comparatively tractable, yet most direction-correct outputs still miss the requested displacement by more than $\pm20\%$, so motion failure is dominated by scale rather than sign. Moreover, executing the camera motion correctly is not sufficient to preserve scene state: systems can reach the right viewpoint yet still lose objects, distort layout, or corrupt appearance. The gap also widens with rollout: all 16 pose-free systems obtain lower SSP on Chain than on Atomic cases, with model-mean drops of $0.164$ for SSP and $0.059$ for CMG. Inverse cycles further expose return-action suppression after self-conditioning, with $66.3\%$ of return steps remaining near-static, indicating that single-step accuracy does not carry over to multi-step consistency. Finally, for the seven multi-image systems, increasing context from $K{=}1$ to $K{=}4$ raises mean CMG from $0.519$ to $0.582$ while SSP decreases from $0.539$ to $0.504$, with outputs frequently attracted toward auxiliary views. Together, these results show that current generators do not reliably couple motion execution with persistent scene state, establishing the need for axis-specific, rollout-aware evaluation.

To test whether the benchmark can also guide model improvement, we follow the same construction principles to build \TrainSet{}, a scene-disjoint resource containing 66,214 trajectories and 108,213 teacher-forced edit pairs. As a diagnostic probe, pairwise SFT raises Qwen-Image-Edit's Overall from $0.495$ to $0.681$, but the gains are highly asymmetric: $+0.303$ in CMG versus $+0.069$ in SSP. Matched controls further show that this transfer is backbone-dependent: on Qwen, SFT reliably improves CMG but yields no confirmed SSP gain, whereas on OmniGen2 the same recipe fails to improve CMG and even reduces SSP---so pairwise supervision does not transfer uniformly across models. Crucially, this asymmetry persists at the full training pool and the longest budget---the most data and training this resource provides---and even on the easiest single-step transitions, where a perfect previous frame precludes error accumulation; it is therefore the pairwise teacher-forced objective, not a shortage of examples, that binds scene preservation. By making this objective-level bottleneck measurable, \Bench{} motivates a shift from single-step pairs to full-trajectory supervision.

\par
\noindent\textbf{Our contributions are as follows.}

\begin{itemize}
    \item We introduce \Bench{}, a 1,400-case benchmark with 2,360 target views for evaluating physical consistency under ego-motion through atomic, chained, and inverse-cycle generation with controlled visual context.
    \item We decompose physical consistency into camera-motion grounding and target-view scene-state preservation, operationalized by CMG-Score and SSP-Score and validated against blinded human judgments.
    \item We benchmark 16 pose-free generators together with two pose-conditioned references, revealing a motion--state gap obscured by visual quality, one-step success, additional visual context, or a single aggregate score.
    \item Using \TrainSet{}, a scene-disjoint set built from the same pipeline, we run controlled SFT studies as a diagnostic probe: pairwise teacher-forced supervision yields asymmetric, backbone-dependent CMG--SSP responses and does not reliably improve both axes, with the asymmetry persisting at the full pool and longest budget---isolating the pairwise teacher-forced objective as the binding constraint---and motivating self-conditioned trajectory supervision.
\end{itemize}

\begin{table}[t]
\centering
{\small
\setlength{\tabcolsep}{2.0pt}
\begin{tabularx}{\columnwidth}{@{}Xcccc@{}}
\toprule
\textbf{Evaluation setting} & \textbf{Ctx.} & \textbf{NL cam.} & \textbf{Multi} & \textbf{State} \\
\midrule
T2I composition~\citep{ghosh2023geneval} & \xmark & \xmark & \xmark & \pmark \\
Spatial gen./editing~\citep{wang2026placebenchmarkingspatialintelligence,xiao2026spatialeditbenchmarkingfinegrainedimage} & \pmark & \pmark & \xmark & \pmark \\
Pose-conditioned NVS/control~\citep{he2025cameractrl} & \cmark & \xmark & \pmark & \pmark \\
Camera/world evaluation~\citep{duan2025worldscore,ying2026wbenchcomprehensivemultiturnbenchmark} & \cmark & \pmark & \cmark & \pmark \\
\midrule
\textbf{\Bench{} (ours)} & \cmark & \cmark & \cmark & \cmark \\
\bottomrule
\end{tabularx}
}
\caption{Scope comparison across six representative benchmarks grouped into four evaluation settings. Columns indicate support for visual context (Ctx.), natural-language camera control (NL cam.), multi-step evaluation (Multi), and explicit cross-view scene-state scoring (State). \cmark/\xmark/\pmark{} denote full, absent, and partial coverage, respectively.}
\label{tab:benchmark_scope}
\end{table}

\section{Related Work}

\paragraph{Image-generation and geometric-consistency benchmarks.}
Prior image-generation evaluation has largely focused on compositional prompt following, spatial relations, and single-output editing~\citep{ghosh2023geneval,wang2026placebenchmarkingspatialintelligence,xiao2026spatialeditbenchmarkingfinegrainedimage}. PDI-Bench extends this focus to generated videos by measuring projective-geometry residuals for scale--depth alignment, 3D motion consistency, and structural rigidity~\citep{wu2026quantitativevideoworldmodel}. \Bench{} complements these settings by jointly scoring language-instructed camera motion and target-view object state across short rollouts.

\paragraph{Novel-view synthesis and camera-controlled generation.}
Novel-view synthesis and 3D reconstruction typically assume calibrated multi-view observations with known or recoverable poses~\citep{reizenstein2021common,dai2017scannet,chang2017matterport3d}. Related camera-controlled generation methods and world-model benchmarks likewise condition on or evaluate explicit camera trajectories~\citep{he2025cameractrl,duan2025worldscore}. \Bench{} instead evaluates pose-free generators driven only by magnitude-specified natural-language camera instructions.

\paragraph{World models and embodied evaluation.}
World models use action-conditioned visual imagination for reasoning and control~\citep{ha2018recurrent,hafner2024masteringdiversedomainsworld,yang2026mindjourney,zhu2026thinkingimaginationagenticvisual}. Recent benchmarks evaluate multi-turn controllability, structural consistency, interaction dynamics, or simulator-grounded execution~\citep{ying2026wbenchcomprehensivemultiturnbenchmark,bagchi2026walkpaintingsegocentricworld,li2026egocentricworldmodelphotorealistic,liu2026kinebenchbenchmarkingembodiedworld}. Unlike these families, \Bench{} evaluates pose-free, language-conditioned camera motion and target-view object-state preservation in static indoor scenes; Table~\ref{tab:benchmark_scope} summarizes these differences. Interaction dynamics and simulator task success remain out of scope.

\section{\Bench{}: Benchmark Design and Construction}
\label{sec:benchmark}

\begin{figure*}[t!]
\centering
\includegraphics[width=\textwidth]{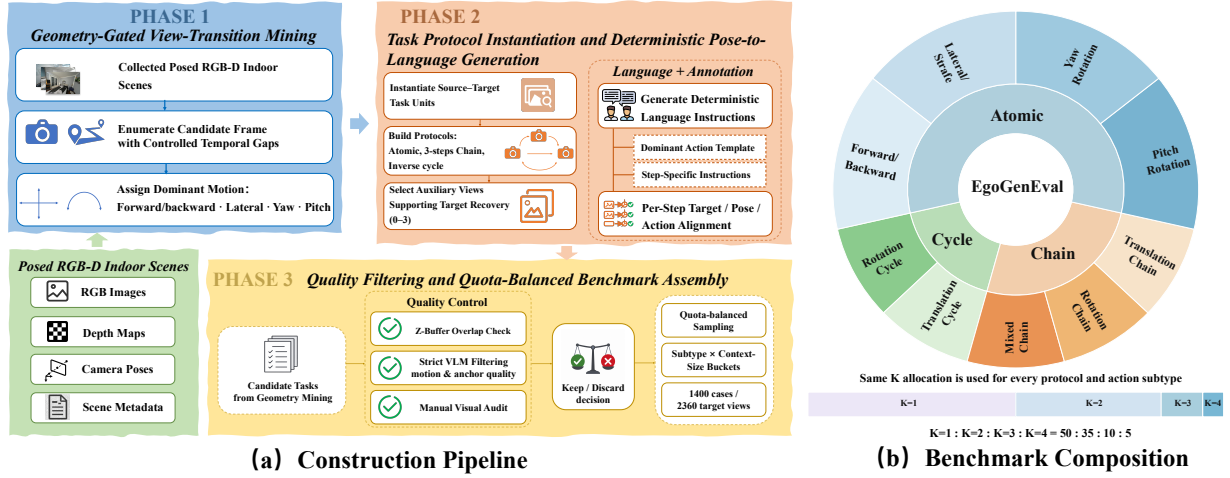}
\caption{Benchmark construction and composition. (a) The construction pipeline mines geometry-grounded view transitions, instantiates language-conditioned atomic/chain/cycle protocols, and applies quality filtering and quota-balanced sampling. (b) The benchmark covers four atomic, three chain, and two cycle subtypes, each allocating $K{=}1,2,3,4$ cases in a $50{:}35{:}10{:}5$ ratio.}
\label{fig:pipeline_composition}
\end{figure*}

\subsection{Task, Context, and Motion Taxonomy}
\label{sec:task_context}

\paragraph{Task formulation.}
Given a current view $I_s$, a natural-language camera-motion instruction $T$ specifying the motion direction and an approximate numerical magnitude (e.g., ``move forward by 0.3\,m''), and optional same-scene auxiliary views $C$, the model generates a target view $\hat{I}_t$:
\begin{equation}
    (I_s,\, C,\, T) \;\rightarrow\; \hat{I}_t.
    \label{eq:task}
\end{equation}

\paragraph{Visual context.}
We vary the total number of input images $K\in\{1,2,3,4\}$: $K{=}1$ uses $I_s$ alone, whereas $K{>}1$ adds $K{-}1$ same-scene auxiliary views. The current view is always placed first, and auxiliary views provide broader scene evidence without coinciding with any evaluated target.

\paragraph{Actions and protocols.}
We retain real-scene transitions dominated by four components: forward/backward translation ($z$), lateral translation ($x$), yaw, and pitch. Because real trajectories contain residual 6-DoF motion, instructions and human judgments consistently evaluate only the specified dominant component. We define three protocols: \textbf{atomic} $f_0{\rightarrow}f_1$, autoregressive three-step \textbf{chain} $f_0{\rightarrow}f_1{\rightarrow}f_2{\rightarrow}f_3$, and inverse \textbf{cycle} $f_0{\rightarrow}f_1{\rightarrow}f_0$. Each transition is scored, yielding three targets per chain and two per cycle.

\subsection{Benchmark Construction}
\label{sec:benchmark_construction}

\Bench{} is constructed in three phases (Figure~\ref{fig:pipeline_composition}(a)).
\emph{Phase 1} mines source--target view pairs with controlled gaps from ScanNet++, ScanNet, HyperSim, and Matterport3D~\citep{yeshwanth2023scannet++,dai2017scannet,roberts2021hypersim,chang2017matterport3d}, computes source-frame relative 6-DoF motion, and retains transitions with a clear dominant forward/backward translation, lateral translation, yaw, or pitch that satisfy overlap and visibility constraints.
\emph{Phase 2} forms atomic, three-step chain, and inverse-cycle cases; each GT relative pose determines the motion, direction, and approximate magnitude, while a frozen overlap-, pose-, and novelty-based ranking selects $K{-}1$ auxiliary views to support target recovery, without using the ground-truth target views as input.
\emph{Phase 3} combines z-buffer checks, VLM~\citep{google2026gemini31propreview} and manual quality filtering, and quota-balanced sampling over subtype and context size, yielding 1{,}400 cases and 2{,}360 target views.
Appendix~\ref{app:construction} details the mining gates, task assembly, and quota-balanced selection of all three phases.

The final benchmark contains 1,400 cases from 771 scenes (at most seven per scene) and 2,360 scored target views; Figure~\ref{fig:pipeline_composition}(b) shows their subtype and context-size distribution. Each unique GT target is annotated once with Qwen3-VL~\citep{bai2025qwen3vltechnicalreport}; the resulting labels are then frozen and unioned with a fixed 36-class indoor vocabulary. The same per-step prompt union is used for GT and generated-image detection and is never exposed to evaluated models.

\section{Evaluation Setup and Metrics}
\label{sec:evaluation}

We operationalize physical consistency in its most basic, static-scene form---\emph{geometric consistency under ego-motion}: whether a generated image realizes the requested camera motion while remaining a valid observation of the same, unchanged 3D environment. Dynamics and lighting are out of scope; this static form is a prerequisite that current generators already fail. We separate spatial fidelity into \emph{camera-motion fidelity}, measured by CMG-Score, and \emph{environment fidelity}, measured by SSP-Score. A realistic image may satisfy either requirement without satisfying the other.

\subsection{Evaluation Setup}
\label{sec:setup}

We evaluate 16 pose-free systems spanning image, multimodal, and video models. \emph{Pose-free} describes only their interface: models receive images and a language instruction but no pose/depth; construction and scoring remain geometry-grounded. Two trajectory-conditioned world models serve as references. Each Chain/Cycle step is a fresh call under a shared wrapper: the first input is the anchor image at step 1 and the preceding output thereafter, followed by retained auxiliaries. For video models, the final frame is scored and becomes the next step's first input. At $K{=}4$, three max-3 systems retain the two highest-overlap auxiliaries; single-reference systems use their native path. Full details are provided in Appendix~\ref{app:interfaces}.

\subsection{CMG: Is the Requested Motion Realized?}
\label{sec:cmg}

CMG measures whether the generated transition realizes the requested ego-motion. For each step $m$, DA3Nested-Giant-Large~\citep{lin2025depth3recoveringvisual} estimates relative motion for both the physical GT transition and the evaluated transition. The latter is source-to-generation for the first step and previous-generation-to-current-generation thereafter. 

Translation is calibrated \emph{per target transition}, not per scene or model:
\begin{equation}
\alpha_m=
\frac{\lVert\mathbf t_m^*\rVert_2}
{\lVert\hat{\mathbf t}^{\mathrm{GT}}_m\rVert_2},
\qquad
\hat{\mathbf t}_m=
\alpha_m\hat{\mathbf t}^{\mathrm{raw}}_m.
\label{eq:cmg_scale}
\end{equation}
Each transition uses the same $\alpha_m$ for all systems, leaving direction and rotation unchanged. Translation magnitude remains target-geometry-assisted, preserving under- or over-scaling only when the estimator scale is consistent across physical and generated pairs.

Let $a_m^*$ be the signed GT value of the instructed dominant component, $\hat a_m$ its estimate, and $d_m=1$ if their signs agree and $0$ otherwise. We jointly score direction and magnitude as
\begin{equation}
q_m=\left(1+\frac{|\hat a_m-a_m^*|}
{\max(|a_m^*|,\epsilon_a)}\right)^{-1}.
\label{eq:cmg_magnitude}
\end{equation}
\begin{equation}
g_m=\tfrac12 d_m(1+q_m).
\label{eq:cmg_step}
\end{equation}
Since $d_m$, $q_m$, and $g_m$ lie in $[0,1]$ with higher values better, we directly average steps within a case:
\begin{equation}
\mathrm{CMG}_c=\frac{1}{T_c}\sum_{m=1}^{T_c}g_m.
\label{eq:cmg_compact}
\end{equation}

where $\epsilon_a=0.1$\,m for translation and $5^\circ$ for rotation. All instructed magnitudes are nonzero by construction (translation $\ge0.15$\,m, yaw $\ge8^\circ$, and pitch $\ge6^\circ$); these floors stabilize relative error rather than define direction, and an exactly zero prediction is direction-wrong. CMG isolates the instructed dominant component; supplementary full-pose diagnostics report unintended off-axis drift, which can lower SSP by displacing the viewpoint without necessarily corrupting the scene. Appendix~\ref{app:full_pose} reports these off-axis and full-pose errors.

\subsection{SSP: Is the Environment Preserved?}
\label{sec:ssp}

SSP measures environment consistency at the intended target view. Grounding DINO~\citep{liu2024grounding} detects objects at box/text thresholds $0.25/0.25$ using a fixed 36-class vocabulary augmented with frozen GT-target labels, with the same detection prompt for GT and generation. A Qwen3-VL~\citep{bai2025qwen3vltechnicalreport} \textsc{propose--demote--recover} matcher with a DINOv3~\citep{simeoni2025dinov3} identity guard produces $M$ one-to-one matches between $G$ evaluable GT objects and $P$ filtered detections. Retention is $F_1=2M/(G+P)$, and unmatched GT objects receive zero spatial and integrity credit. GT pose and depth only determine input-supported target objects and physical depth order. Thus, low SSP may reflect scene corruption, viewpoint error, or evaluator error.

For a target pair $(a,b)$, the planar score measures whether its relative image-plane direction is preserved:
\begin{equation}
O^{xy}_{ab}=
\begin{cases}
\max\!\left(0,\operatorname{cos}(\mathbf c_b^*-\mathbf c_a^*,
\hat{\mathbf c}_b-\hat{\mathbf c}_a)\right), & a,b\text{ matched},\\
0, & \text{otherwise},
\end{cases}
\label{eq:ssp_planar}
\end{equation}
where $\mathbf c_i^*$ and $\hat{\mathbf c}_i$ are box centers and
$\operatorname{cos}(\mathbf u,\mathbf v)
=\mathbf u^\top\mathbf v/
(\lVert\mathbf u\rVert_2\lVert\mathbf v\rVert_2)$.
GT pairs with zero displacement are rejected by data validation; a generated displacement below $10^{-8}$ receives zero.

The depth score checks whether the same object remains in front:
\begin{equation}
O^z_{ab}=\mathbb I\!\left[
\begin{array}{c}
a,b\text{ matched with valid, non-tied depths},\\[-1pt]
\operatorname{sgn}(\hat z_a-\hat z_b)
=\operatorname{sgn}(z_a^*-z_b^*)
\end{array}\right].
\label{eq:ssp_depth}
\end{equation}
The score applies only to GT pairs with reliable depth separation and compares order rather than distance, so generated monocular depth need not be metric. A GT pair is depth-eligible only when its relative depth ratio exceeds \(0.03\).

Pairwise $O^{xy}$ and $O^z$ form topology, averaged with diagonal-normalized box-center accuracy to obtain $\mathcal S$. Appearance integrity $\mathcal I$ weights identity, structure, edges, sharpness, color, and shape by $(.30,.20,.15,.12,.13,.10)$. For multi-step cases, $A_c(x)$ equally weights the step mean and worst step. SSP then averages its three coverage-aware pillars:
\begin{equation}
\mathrm{SSP}_c=\tfrac13\left[A_c(F_1)+A_c(\mathcal S)+A_c(\mathcal I)\right].
\label{eq:ssp_compact}
\end{equation}
Exact matching, position normalization, depth eligibility, appearance weights, and thresholds are provided in Appendix~\ref{app:ssp_details}.

\subsection{Aggregation and Auxiliary Image Metrics}
\label{sec:aggregation}


Within each axis, case scores are normalized to $[0,1]$ with higher values better, so we average cases within Atomic, Chain, and Cycle and then macro-average the protocols. A model-independent GT-only mask removes 44 cases with no evaluable target objects at a required step, retaining 1,356/1,400 cases and 2,265/2,360 steps for every system. CMG and SSP retain this common normalized scale and direction, so Overall weights them equally:
\begin{equation}
\mathrm{Overall}=\frac{1}{2}(\mathrm{CMG}+\mathrm{SSP}).
\label{eq:overall}
\end{equation}

Auxiliary \emph{Reference Similarity} (RefSim) aggregates PSNR~\citep{hore2010image} and LPIPS~\citep{zhang2018unreasonable}; \emph{Visual Quality} (VisQual) aggregates NIQE~\citep{mittal2012making}, MUSIQ~\citep{ke2021musiq}, and CLIP-IQA~\citep{wang2022exploring}. Before averaging, each metric is min--max normalized across the fixed 18-system panel and oriented so higher is better. RefSim measures target-view alignment, whereas VisQual measures perceptual realism; both are excluded from Overall and ranking.

\subsection{Metric Robustness and Human Validation}
\label{sec:human_validation}

\paragraph{Human alignment.} To confirm that CMG and SSP rank models reliably, three rubric-trained annotators independently rank six anonymized systems on 340 steps sampled across protocols and context sizes, scoring camera-motion realization and scene-state preservation separately. At the system level, both metrics align well with human judgment, closely tracking the aggregate human ranking (Spearman $\rho{=}0.943$, Kendall $\tau_b{=}0.867$; exact $p{=}0.0167$). A qualitative comparison further shows that the VLM-assisted matcher resolves identity ambiguities and recovers correspondences a DINO-only baseline misses. These results validate the reliability of CMG and SSP as automatic measures of camera-motion grounding and scene-state preservation; full step-level statistics and rubrics are provided in Appendix~\ref{app:human}.

\paragraph{Evaluator robustness.} To test for evaluator-dependency, we re-score the fixed 18-system panel after substituting each learned component of CMG and SSP and correlate every resulting ranking with the default. Replacing the DA3 relative-pose estimator with VGGT preserves the CMG ranking ($\rho{=}0.943$), and SSP is equally stable across depth backends ($\rho{=}1.000$), Grounding DINO box/text thresholds ($\rho{\geq}0.929$), detection vocabularies ($\rho{=}0.976$), and mean-only temporal aggregation ($\rho{=}0.994$). The consistency of these rankings validates the robustness of the benchmark to the choice of evaluator, demonstrating that its relative results are not driven by any single backend or configuration; full ablations are provided in Appendices~\ref{app:robustness} and~\ref{app:benchmark_robustness}.

\section{Experimental Results}
\label{sec:results}

\subsection{Main Results}
\label{sec:main_results}

\begin{table*}[t!]
\centering
{\small
\setlength{\tabcolsep}{3.0pt}
\renewcommand{\arraystretch}{0.92}
\resizebox{\textwidth}{!}{%
\begin{tabular}{@{}lccccccccccc@{}}
\toprule
\multirow{2}{*}{\textbf{Model}} &
\multirow{2}{*}{\textbf{Overall}$\uparrow$} &
\multicolumn{4}{c}{\textbf{CMG}$\uparrow$} &
\multicolumn{4}{c}{\textbf{SSP}$\uparrow$} &
\multirow{2}{*}{\textbf{RefSim}$^{*}\uparrow$} &
\multirow{2}{*}{\textbf{VisQual}$^{*}\uparrow$} \\
\cmidrule(lr){3-6}\cmidrule(lr){7-10}
& & Avg & Atomic & Chain & Cycle & Avg & Atomic & Chain & Cycle & & \\
\midrule

\multicolumn{12}{c}{\textbf{\textit{Pose-Free Track (Primary Evaluation)}}} \\
\midrule
\multicolumn{12}{c}{\textit{Closed-Source Image Generators}} \\
\midrule
GPT-Image-2~\citeyearpar{openai2026gptimage2}
& 0.66 & 0.72 & 0.74 & 0.72 & 0.71 & 0.60 & 0.66 & 0.55 & 0.60 & 0.59 & 0.90 \\
Seedream-5.0~\citeyearpar{volcengine2026seedream5}
& 0.62 & 0.69 & 0.72 & 0.67 & 0.68 & 0.54 & 0.61 & 0.48 & 0.53 & 0.63 & 0.69 \\
Gemini-3-Pro-Image~\citeyearpar{google2025gemini3proimage}
& 0.57 & 0.59 & 0.60 & 0.60 & 0.58 & 0.55 & 0.62 & 0.49 & 0.55 & 0.71 & 0.57 \\
\midrule

\multicolumn{12}{c}{\textit{Open-Source Image Generators}} \\
\midrule
HiDream-O1~\citeyearpar{cai2026hidream}
& 0.52 & 0.48 & 0.57 & 0.44 & 0.44 & 0.56 & 0.61 & 0.50 & 0.57 & 0.60 & 0.08 \\
FLUX.2-dev~\citeyearpar{flux-2-2025}
& 0.50 & 0.45 & 0.56 & 0.36 & 0.42 & 0.55 & 0.61 & 0.49 & 0.56 & 0.60 & 0.68 \\
HunyuanImage-3.0$^{\dagger}$~\citeyearpar{cao2025hunyuanimage}
& 0.50 & 0.46 & 0.52 & 0.41 & 0.45 & 0.53 & 0.62 & 0.41 & 0.57 & 0.52 & 0.58 \\
Qwen-Image-Edit-2511~\citeyearpar{wu2025qwenimagetechnicalreport}
& 0.50 & 0.47 & 0.57 & 0.41 & 0.43 & 0.52 & 0.62 & 0.40 & 0.54 & 0.47 & 0.56 \\
OmniGen2~\citeyearpar{wu2026omnigen2}
& 0.44 & 0.42 & 0.50 & 0.39 & 0.38 & 0.45 & 0.52 & 0.35 & 0.49 & 0.42 & 0.38 \\
Step1X-Edit~\citeyearpar{liu2025step1x}
& 0.44 & 0.30 & 0.32 & 0.31 & 0.27 & 0.57 & 0.60 & 0.49 & 0.63 & 0.91 & 0.35 \\
ACE++~\citeyearpar{mao2025ace++}
& 0.40 & 0.39 & 0.41 & 0.38 & 0.38 & 0.41 & 0.44 & 0.30 & 0.48 & 0.23 & 0.19 \\
FireRed-Image-Edit-1.1$^{\dagger}$~\citeyearpar{superintelligenceteam2026fireredimageedit10technicalreport}
& 0.39 & 0.41 & 0.48 & 0.38 & 0.36 & 0.37 & 0.50 & 0.21 & 0.42 & 0.48 & 0.47 \\
ICEdit~\citeyearpar{zhang2026enabling}
& 0.32 & 0.34 & 0.36 & 0.33 & 0.33 & 0.30 & 0.33 & 0.24 & 0.31 & 0.00 & 0.48 \\
\midrule

\multicolumn{12}{c}{\textit{Unified Multimodal Models}} \\
\midrule
BAGEL-7B-MoT~\citeyearpar{deng2025emerging}
& 0.45 & 0.44 & 0.46 & 0.43 & 0.41 & 0.47 & 0.55 & 0.39 & 0.46 & 0.57 & 0.32 \\
Emu3.5-Image$^{\dagger}$~\citeyearpar{cui2025emu3}
& 0.32 & 0.44 & 0.42 & 0.47 & 0.45 & 0.21 & 0.32 & 0.10 & 0.19 & 0.42 & 0.39 \\
\midrule

\multicolumn{12}{c}{\textit{Generic Video Models}} \\
\midrule
Kling-2.1~\citeyearpar{kuaishou2025kling21}
& 0.60 & 0.74 & 0.75 & 0.72 & 0.76 & 0.47 & 0.56 & 0.32 & 0.51 & 0.52 & 0.50 \\
Seedance-1.0-Pro-Fast~\citeyearpar{gao2025seedance}
& 0.53 & 0.61 & 0.61 & 0.57 & 0.65 & 0.45 & 0.54 & 0.36 & 0.44 & 0.42 & 0.50 \\
\midrule

\multicolumn{12}{c}{\textbf{\textit{Pose-Conditioned Track (Reference Only)}}} \\
\midrule
HY-WorldMirror-2.0~\citeyearpar{hyworld2026hyworld20multimodalworld}
& 0.66 & 0.85 & 0.84 & 0.88 & 0.82 & 0.48 & 0.48 & 0.42 & 0.55 & 0.90 & 0.37 \\
Lingbot-World~\citeyearpar{robbyantteam2026advancingopensourceworldmodels}
& 0.64 & 0.75 & 0.77 & 0.74 & 0.74 & 0.53 & 0.60 & 0.44 & 0.55 & 0.80 & 0.77 \\
\midrule
\multicolumn{12}{c}{\textbf{\textit{Unranked Score Calibration}}} \\
\midrule
GT-target oracle & 0.94 & 0.98 & 0.98 & 0.98 & 0.98 & 0.90 & 0.92 & 0.87 & 0.90 & -- & -- \\
\bottomrule
\end{tabular}
}
}
\caption{Main leaderboard. Overall is the unweighted mean of the CMG and SSP protocol averages. $^{*}$RefSim and VisQual are normalized auxiliary aggregates and do not affect Overall or ranking. Pose-conditioned systems and calibration rows are reported as references only; the GT-target oracle is a practical evaluator ceiling rather than a mathematical upper bound. $^{\dagger}$These models support at most three total inputs; Appendix~\ref{app:interfaces} describes their $K{=}4$ interface. Appendix~\ref{app:references} provides per-protocol breakdowns.}
\label{tab:main_results}
\end{table*}

\begin{figure*}[tb]
\centering
\captionsetup{hypcap=false}
\includegraphics[width=\columnwidth]{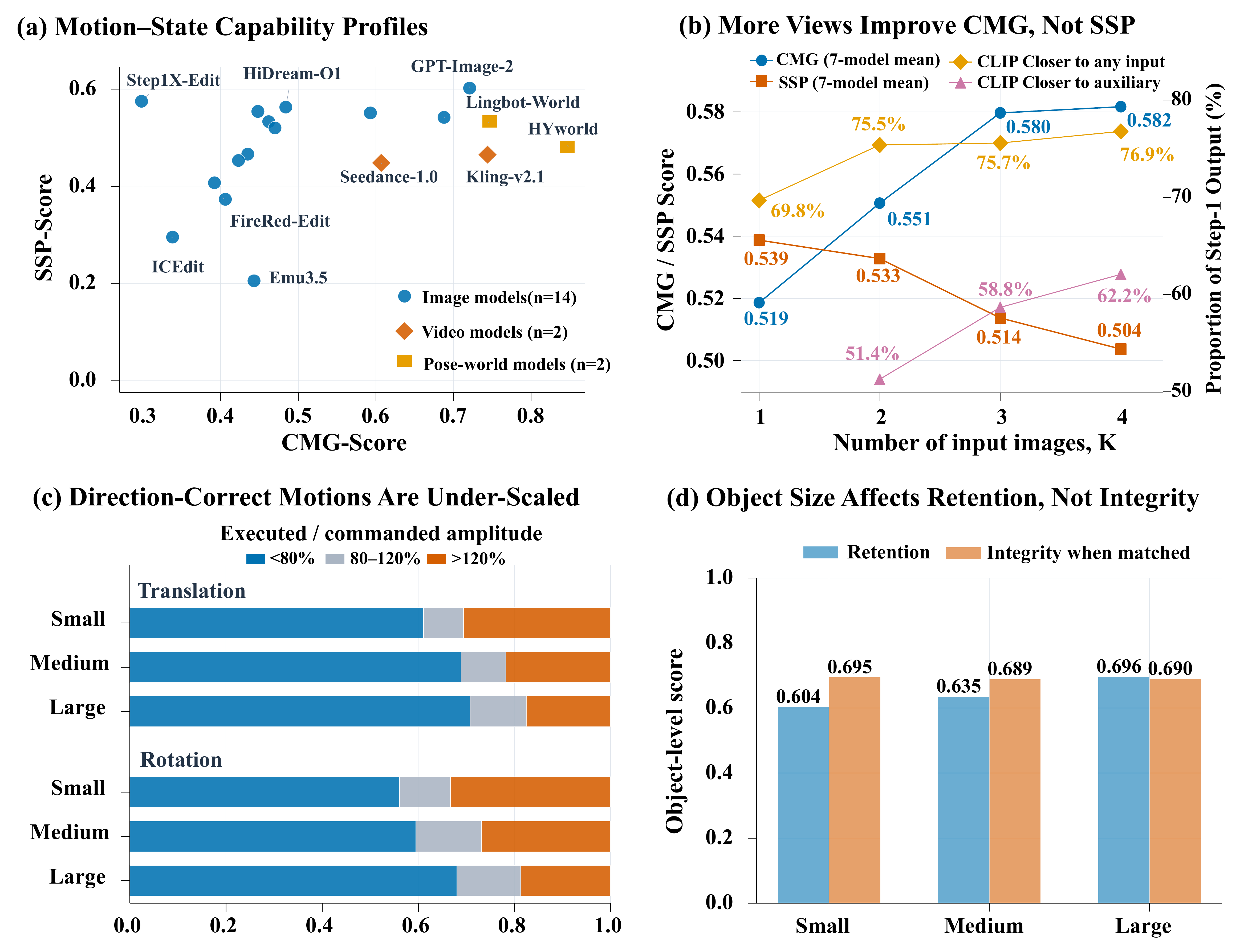}
\caption{Capability profiles and diagnostic results. (a)~CMG--SSP profiles separate motion execution from scene preservation across 18 systems. (b)~For systems supporting all four context settings, additional views improve CMG on average but reduce SSP and increase feature-space attraction toward input views. (c)~Directionally correct outputs commonly under-execute the requested magnitude. (d)~Recall increases with projected object size, whereas integrity among successfully matched objects remains nearly constant. These diagnostics show why visual quality, direction, additional visual context, or matched-object integrity alone is insufficient to establish physical consistency.}
\label{fig:mechanistic_summary}
\end{figure*}

\begin{figure*}[tb]
    \centering
    \includegraphics[width=\textwidth]{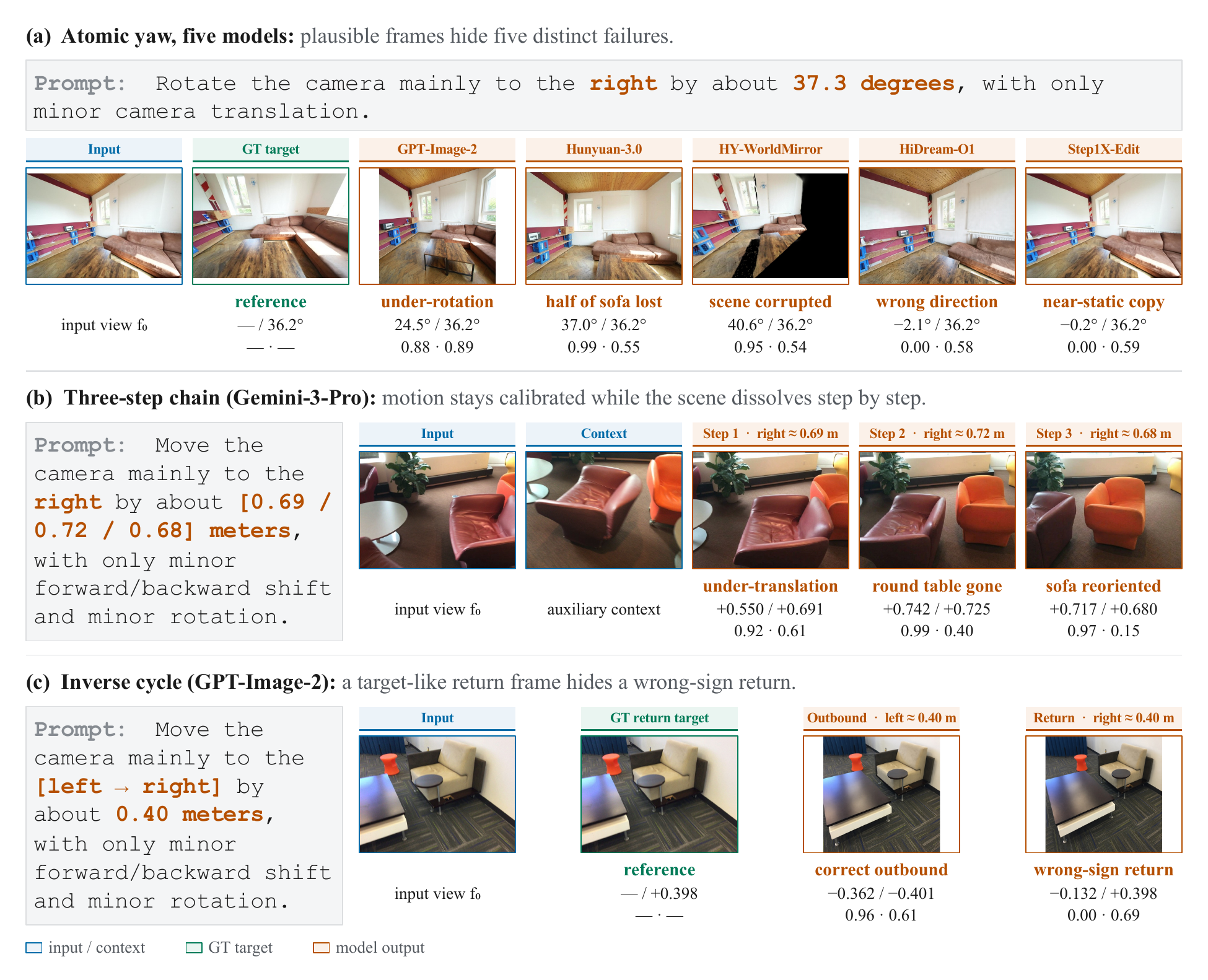}
    \caption{\textbf{Representative failure cases.}
    (a) One atomic yaw instruction given to five models: every frame is
    individually plausible, yet each fails differently---under-rotation,
    object loss, visual corruption, wrong direction, and near-static copying.
    (b) A three-step chain (Gemini-3-Pro): after an under-translated first step
    the motion stays well calibrated (CMG $\ge 0.97$) while the scene dissolves,
    SSP falling $0.61 \rightarrow 0.40 \rightarrow 0.15$.
    (c) An inverse cycle (GPT-Image-2): a correct outbound step is followed by a
    return frame that resembles the target but moves the wrong way
    ($-0.13$\,m against a target of $+0.40$\,m, CMG $0.000$).
    Under each frame we report the predicted and ground-truth motion along the
    instructed axis (degrees or metres), then CMG and SSP.
    Each prompt reproduces the motion clause verbatim and elides the shared
    scaffold: every instruction additionally ends with the scene-preservation
    clause, and multi-image cases are further wrapped with the auxiliary-view
    preamble and the instruction not to copy any input image.
    Bracketed slots mark the single token that differs between the steps of a
    chain or cycle---the magnitude in (b), the direction word in (c).
    Appendix~\ref{app:instruction_template} gives the complete template and all
    eight motion clauses.}
    \label{fig:model_cases}
\end{figure*}

\noindent\textbf{Overall findings of \Bench{}.} Table~\ref{tab:main_results} reports CMG, SSP, and auxiliary metrics for all 18 systems: no system is strong on both axes, the best pose-free Overall ($0.662$) trails the GT-target oracle ($0.940$), and scores fall further from Atomic to Chain and Cycle. The diagnostics below trace these gaps to failures in getting the motion magnitude right and keeping the scene intact, which visual quality does not reveal; Figure~\ref{fig:model_cases} shows representative cases, one per protocol, chosen to illustrate these failures rather than to enumerate every error type.
\begin{itemize}

  \item \textbf{\textit{Motion and scene fidelity are distinct axes: neither score implies the other.}}
  The CMG--SSP scatter in Figure~\ref{fig:mechanistic_summary}(a) shows that a high score on one axis does not imply a high score on the other. HY-WorldMirror attains the highest CMG (\(0.847\)) yet only \(0.481\) SSP, below eight pose-free systems; conversely, FLUX.2-dev reaches \(0.554\) SSP while its CMG is only \(0.448\). The two axes are largely decoupled across the panel, so a single scalar average like Overall can hide which capability a system actually lacks---motion execution and scene preservation must be read separately.

  \item \textbf{\textit{Magnitude failure is systematic, not random.}}
  Getting the \emph{direction} of motion right is the comparatively tractable part of camera-motion grounding: direction accuracy reaches \(0.663\), well above the \(0.525\) obtained by always predicting the single most frequent direction, and ranking systems by direction accuracy alone almost perfectly reproduces their ranking by full CMG (\(\rho = 0.997\))---so what separates one system's CMG from another's is mostly direction, not magnitude. Magnitude, by contrast, is a weakness shared across the board: among direction-correct outputs, \(59.5\%\)--\(70.7\%\) fall outside a \(\pm 20\%\) window around the requested displacement, and this shortfall worsens monotonically as the requested motion grows (\(61.1\%{\to}68.9\%{\to}70.7\%\) for small, medium, and large displacements; Figure~\ref{fig:mechanistic_summary}(c)). Models thus learn the sign of motion but not its magnitude, uniformly and regardless of the scale requested. Privileged conditioning does not remove the error either---it only flips its sign: HY-WorldMirror attains the smallest angular error yet the largest full-translation error, from systematic \emph{over}-scaling rather than under-execution.

  \item \textbf{\textit{Multi-step consistency is not inherited from single-step execution.}}
  All 16 pose-free systems lose SSP from Atomic to Chain (mean drop \(0.164\)), and the degradation spans all three SSP pillars with comparable declines for large furniture and small portable objects (\(0.195\) vs.\ \(0.191\)). Inverse-Cycle rollout reveals a complementary failure: within a single cycle, \(66.3\%\) of return steps are near-static, versus only \(37.4\%\) on the outbound step. Appendix~\ref{app:cycle_behavior} defines the near-static criterion and reports the per-step response ratios and direction accuracies behind these rates. Because both are single-step actions of comparable difficulty, this near-doubling of the failure rate isolates self-conditioning, not single-step ability, as the cause: the return step must condition on a self-generated view, and that is where execution collapses. Rollout consistency is therefore a separate capability, not a downstream consequence of single-step quality, and motivates training the model on its own generated intermediate views, the inputs it must actually condition on at inference, rather than only on pairs of adjacent real frames.

  \item \textbf{\textit{Additional context views are imitated, not fused.}}
  For seven multi-image models, increasing context from \(K{=}1\) to \(K{=}4\) raises mean CMG from \(0.519\) to \(0.582\), while SSP decreases from \(0.539\) to \(0.504\) (Figure~\ref{fig:mechanistic_summary}(b)). To rule out that this drop merely reflects a change in which objects are scored as \(K\) grows, we fix the evaluated object set to those objects present at every \(K\) and recompute SSP: the decline persists on this fixed set (\(\Delta = -0.035\) to \(-0.037\)). At \(K{=}4\), \(62.2\%\) of outputs are CLIP-closest to an auxiliary view rather than the current target view, while only \(0.6\%\) are near-pixel copies; auxiliary-confused rows have systematically lower SSP (\(\Delta = -0.064\), 95\% CI \([-0.119, -0.009]\)). The additional views carry genuine evidence about the scene, yet current models cannot exploit it as such: rather than fusing these views into a more consistent target, they merely imitate them. More context is therefore not a monotone path to physical consistency.

  \item \textbf{\textit{Perceptual quality is statistically decoupled from spatial consistency.}}
  Across the 14 image and editing systems with full quality signals, VisQual correlates with SSP at only \(\rho = 0.292\) (\(p = 0.310\), n.s.) and with CMG at \(\rho = 0.582\) (\(p = 0.032\)). The most striking instance: Step1X-Edit holds the highest RefSim in the full panel (\(0.908\)) while recording the lowest pose-free CMG (\(0.298\)): its outputs are near-static, remaining visually close to the target yet failing to execute the requested action. A system that looks polished is statistically indistinguishable from one that is spatially incoherent. Checkpoint selection based on no-reference quality metrics will not surface magnitude under-execution or multi-step scene dissolution; the CMG--SSP frontier is the necessary signal.
\end{itemize}

\paragraph{Correlation with other benchmarks.}
Table~\ref{tab:cross_benchmark} compares system rankings on \Bench{} against a meta-ranking aggregated from four external editing benchmarks. The model rankings on \Bench{} closely align with the meta-rankings, which validates \Bench{} as a reliable measure of editing quality rather than an artifact of our protocol. At the same time, \Bench{} diverges most from the spatial-editing benchmark closest to our setting, SpatialEdit. Existing benchmarks reward appearance preservation without asking whether the requested viewpoint was physically realized, so this divergence shows that \Bench{} captures a capability the closest prior benchmark misses.

\begin{table}[tb]
\centering
{\small
\setlength{\tabcolsep}{0.8mm}
\begin{tabular}{@{}lcccccc@{}}
\toprule
\multirow{2}{*}{\textbf{Model}} &
\textbf{GEdit} &
\textbf{Img} &
\multirow{2}{*}{\textbf{RISE}} &
\textbf{Sp.} &
\textbf{Meta} &
\textbf{Rank} \\
& \textbf{v2} & \textbf{Edit} & &
\textbf{Edit} & \textbf{Rank} & \textbf{(Ours)} \\
\midrule
Qwen-Image-Edit-2511 & 1 & 1 & 1    & 2 & \textbf{1} & \textbf{1} \\
Step1X-Edit          & 2 & 2 & 3    & 1 & \textbf{2} & \textbf{4} \\
BAGEL                 & 4 & 4 & 2    & 4 & \textbf{4} & \textbf{2} \\
OmniGen2              & 3 & 3 & $=4$ & 3 & \textbf{3} & \textbf{3} \\
ICEdit                & 5 & 5 & $=4$ & 5 & \textbf{5} & \textbf{5} \\
\bottomrule
\end{tabular}
}
\caption{Rank comparison with four external benchmarks%
~\citep{jiang2026geditbenchv2humanalignedbenchmark,ye2026imgedit,zhao2026envisioning,xiao2026spatialeditbenchmarkingfinegrainedimage}.
Sp.\ Edit denotes SpatialEdit. Meta-Rank aggregates the four external
rankings; Ours ranks models by \Bench{} Overall, and ``$=$'' denotes
tied ranks.}
\label{tab:cross_benchmark}
\end{table}

\subsection{Supervised Fine-Tuning}
\label{sec:sft}

To further examine the utility of \Bench{} beyond evaluation, this section investigates whether its construction pipeline can produce a supervised fine-tuning (SFT) dataset, \TrainSet{}, for improving existing generators, from which we draw several observations.

\noindent\textbf{\TrainSet{} construction.} \TrainSet{} uses the same geometry-grounded pipeline as \Bench{}, but it is scene-disjoint: when assembling \TrainSet{}, we exclude every scene used in \Bench{}, so all reported gains are measured on scenes never seen during fine-tuning. Candidate trajectories first pass the same motion-dominance, overlap, visibility, depth-layer, and pose-separation gates as the benchmark. A curation stage then drops trajectories with ambiguous motion, unusable images, cross-scene sequences, dominant dynamic content, or no trackable spatial anchors, and a bidirectional depth-consistency check keeps only trajectories whose cross-view overlap stays within \([0.35, 0.85]\) at every step. The resulting pool has 66,214 trajectories and 108,213 teacher-forced transition pairs. We form the pairs by taking each multi-step trajectory and splitting it into its adjacent frames $f_{i-1}\!\to\!f_i$, then training each pair on its own with the ground-truth previous frame as context.

\noindent\textbf{Training details.} We fine-tune Qwen-Image-Edit-2511 with rank-16 LoRA on this pool. As a cross-backbone control, we also fine-tune OmniGen2 on the same pool against a matched base run. All variants are evaluated on the same 1,400 held-out cases. Appendix~\ref{app:sft} reports the full provenance, budgets, and confidence intervals.

\begin{figure*}[t!]
\centering
\includegraphics[width=\textwidth]{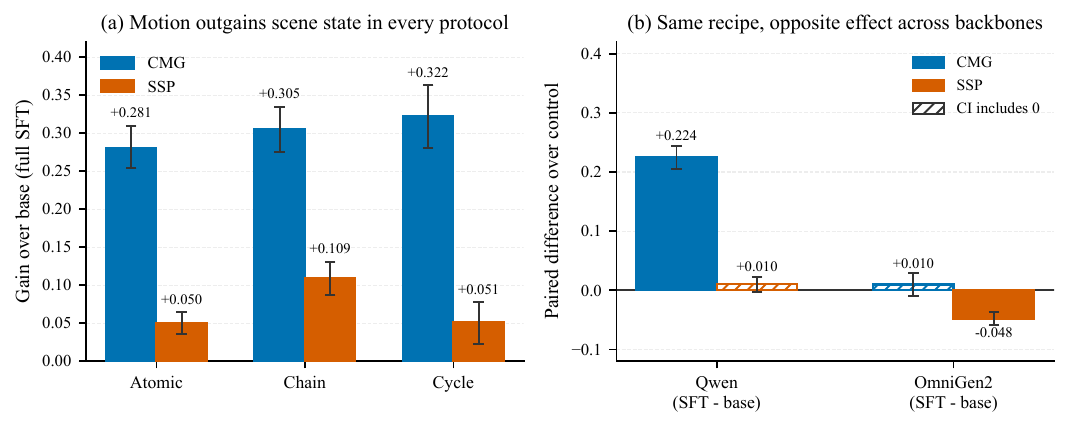}
\caption{SFT effects on the two axes. \textbf{(a) More training moves motion, not scene state.} On the full cleaned pool, deltas relative to the Qwen base as the budget grows (base\(\to4\text{k}\to13.5\text{k}\) updates): CMG climbs steeply (\(+0.189\to+0.303\)) while SSP crawls up from near zero (\(+0.015\to+0.069\)). This is the most data and training the resource provides, yet the axes stay far apart. \textbf{(b) Same recipe, opposite effect across backbones.} Paired deltas ($\Delta$CMG, $\Delta$SSP) relative to each model's base at a matched 4k-update budget on the same cleaned pool, under 20,000 scene-cluster bootstrap resamples: Qwen SFT gains CMG strongly but its SSP change is indistinguishable from zero, while OmniGen2 SFT gains nothing on CMG and loses SSP. Hatched bars mark deltas whose $95\%$ interval includes zero. At matched compute, neither backbone gains scene preservation. Intervals quantify benchmark sampling uncertainty, not training-run variance.}
\label{fig:sft_summary}
\end{figure*}

\noindent\textbf{SFT results.} Full SFT raises Qwen Overall from \(0.495\) to \(0.681\), above the best off-the-shelf pose-free model at \(0.662\), and Chain SSP rises from \(0.397\) to \(0.505\). But the two axes move very unevenly: CMG rises by \(+0.303\) while SSP rises by only \(+0.069\). The gap is not simply a matter of training longer (Figure~\ref{fig:sft_summary}a): at a \(4\text{k}\)-update budget SSP has barely moved (\(+0.015\)), and tripling the budget to \(13.5\text{k}\) is what lifts it to \(+0.069\)---while CMG has already gained \(+0.189\) by \(4\text{k}\). Scene preservation does improve, but slowly and always far behind motion: fine-tuning readily teaches the model where to move, not how to hold the scene together while moving.

\noindent\textbf{Cross-backbone transfer.} The same recipe does not carry over to a second model. Comparing both backbones at a matched 4k-update budget on the same cleaned pool: on Qwen, fine-tuning improves CMG but not SSP; on OmniGen2 (Figure~\ref{fig:sft_summary}b), it does the opposite, gaining nothing on CMG while losing scene state across object retention, spatial relations, and appearance. On one model the transition-pair recipe helps motion, on another it hurts the scene. The direction of the CMG effect is thus backbone-dependent, but the two backbones agree on the axis that matters here: at matched compute, neither gains scene preservation.

\noindent\textbf{The bottleneck is the objective, not the data.} The asymmetry survives every more-favorable condition this resource allows. It holds at the full \(108\text{k}\)-pair pool trained for the longest budget---the most data and training available---where SSP still trails CMG several times over. It holds when the data is made cleaner: at a matched \(4\text{k}\)-update budget and matched pool size, quality-filtering the \(40\text{k}\) pool raises CMG by \(+0.045\) but SSP by only \(+0.003\) over the unfiltered pool. And it holds on the easiest cases: on single-step Atomic transitions, where the model is handed a perfect previous frame and no error can accumulate, SSP reaches only \(0.669\) (a \(+0.051\) gain) while the matched CMG gain is several times larger. When more data, cleaner data, and the easiest error-free version of the task all leave scene preservation this far behind, the ceiling is set by what the training signal rewards: pairwise next-view prediction scores getting the viewpoint right far more than keeping objects, their relations, and their appearance intact. We therefore read this as a limit of the pairwise SFT objective, not a call for a larger or better-curated dataset.

\noindent\textbf{Toward better supervision.} If the objective is the binding constraint, the remedy is to change what the objective supervises. Scene state can be supervised directly, using the pose, depth, visibility, and cross-step object correspondences that \TrainSet{} already carries but plain SFT ignores, so that preservation becomes an explicit target instead of something the model must infer from next-view prediction alone. Training on the model's own multi-step rollouts, rather than always on ground-truth frames, would in addition build robustness to the errors that accumulate over a sequence.

\section{Discussion and Limitations}
\label{sec:discussion}

\paragraph{A persistent-state bottleneck.}
Across the benchmark, current systems often generate the \emph{appearance} of a camera action without maintaining the persistent scene state on which that action operates: direction can be predicted without metric control, auxiliary views are imitated rather than integrated, and a self-generated view is a weak anchor for the inverse action. Small, portable objects are also retained less often than large furniture. These observations do not establish a particular internal representation, but they localize the difficulty to coupling camera transformation with cross-view scene state, and the SFT study (Section~\ref{sec:sft}) shows the same gap persists under pairwise supervision. Overall remains a useful summary, yet close scores should be read alongside CMG--SSP profiles and protocol-level uncertainty.

\paragraph{Limitations.}
\Bench{} covers static indoor scenes, four camera-action families, and rollouts of up to three steps; dynamic scenes, wider motions, and longer horizons remain open. Because CMG and SSP are computed from learned perception models, a low score can reflect the limits of these evaluators rather than a true failure of the generator, even though both metrics track blinded human rankings. Appendix~\ref{app:limitations} details the statistical treatment, evaluator, licensing, interface, and contamination boundaries.

\FloatBarrier

\section{Conclusion}
\label{sec:conclusion}

We introduced \Bench{}, a 1,400-case benchmark of physical consistency under ego-motion, measuring camera-motion and target-view environment fidelity with metrics validated against aggregate human system rankings. Across 16 pose-free systems, similar Overall scores conceal distinct CMG--SSP profiles, scene state degrades consistently in Chain rollouts, and additional views improve motion grounding without improving preservation---together establishing \Bench{} as an axis-specific, rollout-aware diagnostic for physically consistent generation under ego-motion. As a controlled study of whether benchmark-derived data can narrow this gap, we built \TrainSet{} from the same pipeline and ran matched SFT experiments across backbones. Pairwise teacher-forced supervision improves camera-motion grounding but does not reliably improve scene-state preservation, its effect does not transfer uniformly across backbones, and the asymmetry persists at the full pool and the longest budget and even on the easiest single-step cases---isolating the pairwise teacher-forced objective as the binding constraint. Together, the evaluation and training results identify persistent scene state as a central open challenge and motivate self-conditioned trajectory training with explicit pose and visibility supervision.

\bibliographystyle{unsrtnat}
\bibliography{main}

\beginappendix
\section{Appendix Roadmap and Evaluation Scope}
\label{app:overview}

This appendix is included in the same arXiv document so that the complete benchmark contract can be read and cited together with the main results.

The primary evaluation comprises 16 pose-free generators; HY-WorldMirror-2.0 and Lingbot-World receive explicit 6-DoF trajectories and are reported separately as pose-conditioned references, yielding 18 evaluated systems in total. Unless stated otherwise, diagnostic conclusions use the 16-system primary scope, and pose-conditioned references are italicized in appendix tables rather than ranked.

\section{Benchmark Construction}
\label{app:construction}

\subsection{Source data and candidate mining}

\Bench{} draws posed indoor RGB-D observations from ScanNet++~\citep{yeshwanth2023scannet++}, ScanNet~\citep{dai2017scannet}, HyperSim~\citep{roberts2021hypersim}, and Matterport3D~\citep{chang2017matterport3d}. The final 1{,}400 cases contain 1{,}003, 210, 165, and 22 cases from these sources, respectively. These counts are the outcome of feasibility-constrained balanced selection rather than a target population estimate.

For a candidate transition from frame $i$ to frame $j$, we compute the relative six-degree-of-freedom (6-DoF) transform in the source-camera frame,
\begin{equation}
\Delta P_{i\rightarrow j}=(t_x,t_y,t_z,\psi,\theta,\phi),
\end{equation}
where $t_x$ and $t_z$ represent lateral and forward/backward translation, while $\psi$ and $\theta$ represent yaw and pitch. Candidate mining removes near-duplicate views, excessive motion, low-overlap pairs, and transitions that fail motion-clarity, visibility, or pose-gap gates. Among legal components, the normalized largest component defines the dominant action. Residual translation, roll, or rotation may remain because the data come from real trajectories; they do not create an additional instruction class or scoring axis.

Translation and rotation components are normalized by fixed minimum-action gates, and the largest legal component determines the action label only after overlap, visibility, and motion-separation checks. Translation candidates span $0.15/0.18$--$1.20$\,m (source-dependent lower bound), yaw $8^\circ$--$45^\circ$, and pitch $6^\circ$--$35^\circ$. Candidate overlap lies in $[0.35,0.85]$, the dominant component must exceed competing components by at least $1.5$, and translation--rotation separation must be at least $1.8$. Atomic and Chain frame gaps are 2--50 and 2--25 frames; pose-NMS uses $0.30$\,m and $8^\circ$ bins. The per-case metadata records the source-specific translation minimum and every gate outcome.

\subsection{Task assembly and instruction construction}

Atomic cases contain one real current-to-target transition. Three-step chains use three valid consecutive transitions. C1 contains three translations from the same atomic family, C2 contains three rotations from the same atomic family, and C3 contains at least two atomic action families. Inverse cycles contain a planned two-step path
\begin{equation}
f_0 \xrightarrow{a} f_1 \xrightarrow{-a} f_0.
\end{equation}
The return action is computed from the inverse GT transform in the $f_1$ camera frame. Y1 uses a translational outbound action and Y2 a rotational outbound action. The return is fixed before inference and is not adapted to the model's realized outbound motion.

For every step, the GT relative pose determines the dominant action class, sign, and magnitude-specified value in meters or degrees. A deterministic template adds a preservation instruction covering scene identity, layout, relative depth, and perspective. The full pose, target image, depth, object labels, and detector prompts are never exposed to pose-free models.

The total input count is $K\in\{1,2,3,4\}$. The first image is the current view at step~1 and the previous generated output at later steps. The remaining $K-1$ images are same-scene auxiliary views. They are ranked by target-visible support, overlap with the current/target views, pose distance, and view novelty; an evaluated target is never supplied as context. This target-aware selection is evaluator-side metadata used to choose evidence, not target-image access by the evaluated model.

\subsection{Filtering and quota-balanced selection}

\begin{figure*}[t]
\centering
\includegraphics[width=\textwidth]{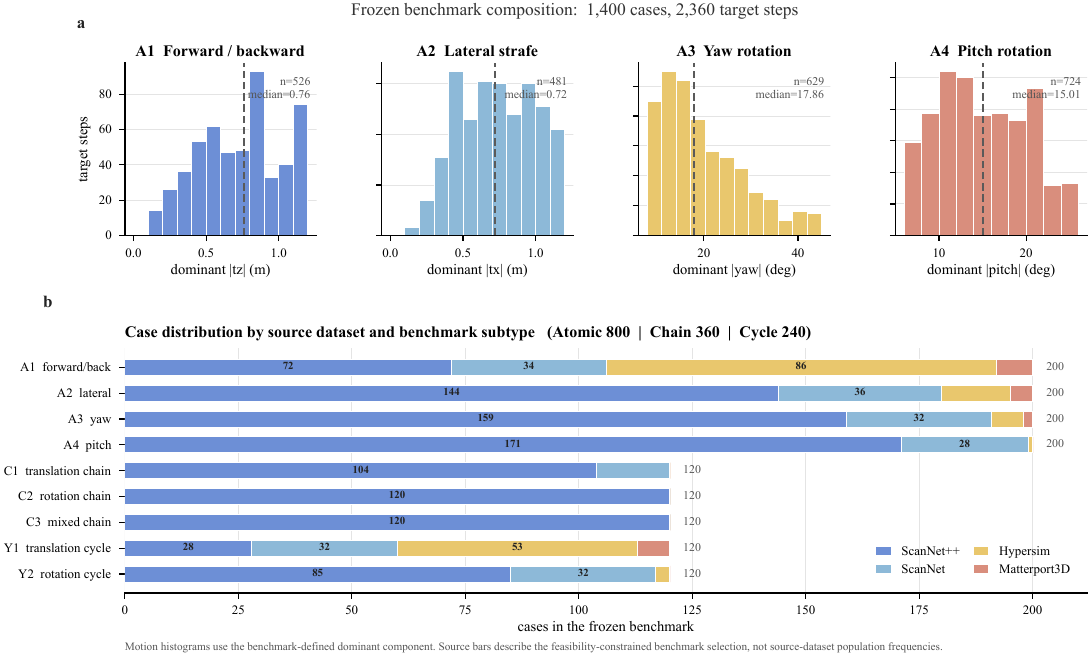}
\caption{Frozen benchmark composition. \textbf{(a)}~Per-family histograms of the benchmark-defined dominant component: forward/backward and lateral translation in meters, yaw and pitch rotation in degrees, each annotated with its step count and median. \textbf{(b)}~Case distribution across the four source datasets within every Atomic, Chain, and Cycle subtype. The bars describe the feasibility-constrained balanced selection actually shipped, not source-dataset population frequencies.}
\label{fig:dataset_stats}
\end{figure*}

The final selection uses 36 subtype-by-$K$ buckets. Each of the four atomic subtypes contributes 200 cases, while each of the three chain and two cycle subtypes contributes 120. Within every subtype, the $K=1,2,3,4$ allocation follows $50{:}35{:}10{:}5$, yielding 700, 490, 140, and 70 cases. Quota-constrained selection balances source datasets and limits scene concentration, producing 771 scenes with at most seven cases per scene. All selected cases pass z-buffer, VLM, and manual visual-quality checks. Figure~\ref{fig:dataset_stats} summarizes the resulting benchmark: the top row reports the dominant-component magnitude distribution for each atomic family, and the bottom row reports how cases divide across the four source datasets within every subtype. The source split reflects feasibility-constrained balanced selection rather than a source-population estimate.

\begin{table}[t]
\centering
\small
\setlength{\tabcolsep}{3pt}
\begin{tabular}{llrrrrrr}
\toprule
\textbf{Protocol} & \textbf{Subtypes} & \textbf{Cases} & \textbf{Steps} & \textbf{$K$1} & \textbf{$K$2} & \textbf{$K$3} & \textbf{$K$4} \\
\midrule
Atomic & A1--A4 & 800 & 800  & 400 & 280 & 80 & 40 \\
Chain  & C1--C3 & 360 & 1080 & 180 & 126 & 36 & 18 \\
Cycle  & Y1--Y2 & 240 & 480  & 120 & 84  & 24 & 12 \\
\midrule
\textbf{Total} & 9 subtypes & \textbf{1400} & \textbf{2360} & \textbf{700} & \textbf{490} & \textbf{140} & \textbf{70} \\
\bottomrule
\end{tabular}
\caption{Benchmark composition. $K$ counts all model inputs, including the current or previous rollout view. Each case has one $K$ setting in the main evaluation; the controlled Context-$K$ study separately repeats matched cases across $K$.}
\label{tab:benchmark_statistics}
\end{table}

\subsection{Examples and data separation}

\begin{figure*}[t]
\centering
\includegraphics[width=\textwidth]{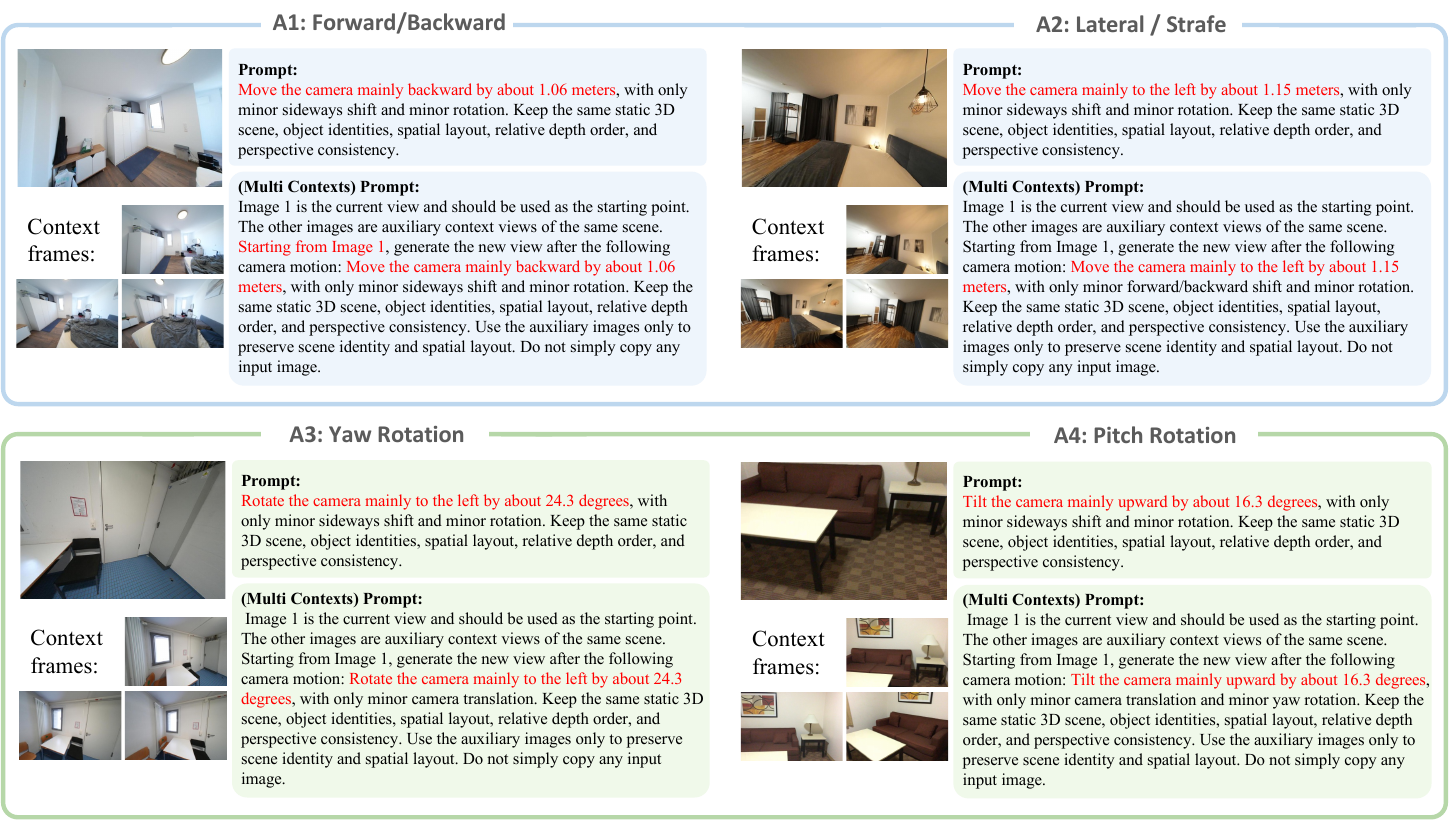}
\caption{Representative benchmark cases for forward/backward translation, lateral translation, yaw, and pitch. Each case contains the current view, optional same-scene context, and a magnitude-specified instruction. Target images and evaluator metadata are not shown to pose-free models.}
\label{fig:task_examples}
\end{figure*}

The planned release will provide versioned case manifests, construction metadata, evaluator contracts, and the inputs required to reproduce the reported aggregate results. For \TrainSet{}, it will additionally provide row-level trajectory metadata, deterministic teacher-forced pair conversion, and the scene-disjointness audit. It will not reproduce every intermediate candidate generated during early data mining, and restricted source pixels will remain governed by the licenses of the underlying datasets. The evaluation scenes are disjoint from the fine-tuning data used in Section~\ref{app:sft}; pretraining exposure of the evaluated foundation models to the source datasets remains unknown.

\section{Model Interfaces and Rollout Protocol}
\label{app:interfaces}

\subsection{Evaluation scope}

\begin{table}[t]
\centering
\small
\setlength{\tabcolsep}{3pt}
\begin{tabularx}{\columnwidth}{@{}p{2.05cm}X@{}}
\toprule
\textbf{Family} & \textbf{Systems} \\
\midrule
Closed image & GPT-Image-2, Seedream-5.0, Gemini-3-Pro-Image \\
\makecell[l]{Open gen./\\edit} & HiDream-O1, HunyuanImage-3.0, Qwen-Image-Edit-2511, FLUX.2-dev, OmniGen2, Step1X-Edit, ACE++, FireRed-Image-Edit-1.1, ICEdit \\
Unified & BAGEL-7B-MoT, Emu3.5-Image \\
\makecell[l]{Pose-free\\video} & Kling-2.1, Seedance-1.0-Pro-Fast \\
Pose-cond.\ ref. & HY-WorldMirror-2.0, Lingbot-World \\
\bottomrule
\end{tabularx}
\caption{Evaluated systems. The first four rows contain the 16-model Pose-Free Track; the final row contains two pose-conditioned references reported separately from the primary ranking.}
\label{tab:model_scope}
\end{table}

Each transition is a fresh model call without conversation history. The rollout is nevertheless autoregressive: the generated output at step $m-1$ replaces the current view at step $m$. Auxiliary views remain available according to the native interface, but the original $f_0$ is not re-added on Cycle step~2. Consequently, a cycle cannot close by receiving the answer view as an extra reference. Every model--step contributes one frozen output; case bootstrap therefore quantifies benchmark-sampling uncertainty, not stochastic generation variance.

Multi-image models receive the current/previous view followed by retained auxiliaries. Step1X-Edit and ICEdit use their single-reference paths; ACE++ uses its single-reference path and builds on ACE and FLUX.1-dev~\citep{mao2025ace++,hanace,flux2024}. Emu3.5, FireRed-Edit, and HunyuanImage-3.0 support at most three total images. On a $K=4$ case, these models retain the current/previous rollout view and the two auxiliaries with highest mean target-visible overlap; original order breaks ties. The controlled Context-$K$ study excludes native max-three and single-reference paths.

The two pose-free video models are invoked once per benchmark step; the final video frame is scored and becomes the next current view. Their table labels carry an ``Autoreg'' suffix that names this rollout wrapper rather than a different model checkpoint. HY-WorldMirror-2.0 and Lingbot-World receive native cumulative 6-DoF trajectories and are scored at the same step boundaries. Their scores are references rather than directly comparable pose-free baselines.

Evaluator preprocessing is component-specific. The official CMG pipeline preserves native aspect ratio; the aspect-ratio sensitivity test in Section~\ref{app:robustness} applies a center crop only as an intervention. SSP uses the frozen detector and depth-estimator preprocessing described in Section~\ref{app:ssp_details}. No blanket resize-to-GT rule defines the benchmark.

\subsection{Prompt templates}
\label{app:prompts}
\label{app:instruction_template}

Every instruction in the benchmark is produced by filling one fixed template with the measured relative camera pose of the step. No instruction is written by hand or by a language model, so the mapping from pose to text is exact and reproducible.

Let $\Delta P_m=P_{m-1}^{-1}P_m$ be the GT relative transform for step $m$, and let $a_m^*$ be its dominant scalar component. A deterministic template converts $\operatorname{sgn}(a_m^*)$, $|a_m^*|$, and the unit into a motion clause: four clause templates, one per atomic family, each carrying a binary direction slot, so eight realized clauses in total (Table~\ref{tab:motion_clauses}). The magnitude is $|a_m^*|$ rounded to two decimals for translation (meters; observed range $0.15$--$1.20$) and to one decimal for rotation (degrees; observed range $8.1$--$44.8$ for yaw and $6.1$--$25.0$ for pitch); the direction word is selected by $\operatorname{sgn}(a_m^*)$. Checked against \texttt{pose\_metadata} for all $2{,}360$ instructions, both fill rules hold with zero exceptions.

Let $\langle\textsc{motion}\rangle$ denote that clause. For $K=1$ the instruction is the clause followed by the scene-preservation sentence, with no preamble:

\begin{quote}\ttfamily\small\noindent
$\langle$MOTION$\rangle$ Keep the same static 3D scene, object identities, spatial layout, relative depth order, and perspective consistency.
\end{quote}

\noindent
For $K\ge 2$ the same clause is wrapped so that the current view is unambiguous and the auxiliary views cannot be copied:

\begin{quote}\ttfamily\small\noindent
Image 1 is the current view and should be used as the starting point. The other images are auxiliary context views of the same scene. Starting from Image 1, generate the new view after the following camera motion: $\langle$MOTION$\rangle$ Keep the same static 3D scene, object identities, spatial layout, relative depth order, and perspective consistency. Use the auxiliary images only to preserve scene identity and spatial layout. Do not simply copy any input image.
\end{quote}

\noindent
The current view is always referred to as \texttt{Image~1}, independently of $K$. Multi-step tasks introduce no additional wording: each step of a chain ($3$ steps) or of a cycle ($2$ steps) is an independent instantiation of the same template. The two clauses of a cycle therefore differ only in the direction word ($240/240$ cycles), and within the single-family chains C1 and C2 the three clauses differ only in the magnitude ($240/240$); C3 mixes action families by construction, so its clauses differ in the clause template as well. Blanking the numerals in all $2{,}360$ instructions leaves exactly $16$ distinct forms---two wrappers $\times$ eight motion clauses---with no seventeenth. Figure~\ref{fig:task_examples} shows instantiated single- and multi-image prompts rather than schematic paraphrases.

\begin{table}[t]
\centering
\small
\caption{The eight motion clauses. \textit{Axis} is the component of the relative pose that determines both the magnitude and the direction word; $d$ is in meters and $a$ in degrees.}
\label{tab:motion_clauses}
\begin{tabular}{@{}llp{0.55\textwidth}@{}}
\toprule
Subtype & Axis & Clause \\
\midrule
A1 forward/backward & $t_z$ &
\texttt{Move the camera mainly \{forward\,$|$\,backward\} by about $d$ meters, with only minor sideways shift and minor rotation.} \\[2pt]
A2 lateral strafe & $t_x$ &
\texttt{Move the camera mainly to the \{right\,$|$\,left\} by about $d$ meters, with only minor forward/backward shift and minor rotation.} \\[2pt]
A3 yaw rotation & yaw &
\texttt{Rotate the camera mainly to the \{right\,$|$\,left\} by about $a$ degrees, with only minor camera translation.} \\[2pt]
A4 pitch rotation & pitch &
\texttt{Tilt the camera mainly \{upward\,$|$\,downward\} by about $a$ degrees, with only minor camera translation and minor yaw rotation.} \\
\bottomrule
\end{tabular}
\end{table}

For a chain, $T_1,T_2,T_3$ are independently derived from the three GT transitions:
\begin{equation}
\begin{aligned}
\hat f_1&=M(f_0,C,T_1),\\
\hat f_2&=M(\hat f_1,C,T_2),\\
\hat f_3&=M(\hat f_2,C,T_3).
\end{aligned}
\label{eq:chain_rollout}
\end{equation}
The GT sequence determines each instruction, while the generated sequence determines the visual current view. The model never receives a GT intermediate frame. For a cycle, the return prompt is derived from $\Delta P_{1\rightarrow0}$ rather than from a string-level direction swap.

\section{Complete Evaluation Protocol}
\label{app:metrics}

\subsection{Camera Motion Grounding (CMG)}
\label{app:cmg_details}

CMG uses the camera decoder of DA3Nested-Giant-Large~\citep{lin2025depth3recoveringvisual} at resolution 504 with upper-bound resizing, native aspect ratio, saddle-balanced reference selection, and ray-pose disabled. At step $m$, it estimates the relative motion of the physical GT pair and the evaluated pair. For translation, monocular scale is calibrated from the GT pair,
\begin{equation}
\alpha_m=\frac{\lVert\mathbf t_m^*\rVert_2}
{\lVert\hat{\mathbf t}^{\mathrm{GT}}_m\rVert_2},
\qquad
\hat{\mathbf t}_m=\alpha_m\hat{\mathbf t}^{\mathrm{raw}}_m.
\end{equation}
The same factor is applied to the current-to-generated estimate; it changes neither direction nor rotation. Calibration is performed independently for each physical target transition, never per model, scene, or global dataset. Hence every system evaluated on a transition receives the identical $\alpha_m$. This makes translation magnitude interpretable for that transition but leaves magnitude-dependent conclusions conditional on the learned pose backend and GT-assisted calibration.

Let $a_m^*$ and $\hat a_m$ be the GT and estimated signed values of the instructed component. Direction and magnitude agreement are
\begin{equation}
\begin{aligned}
d_m&=\mathbb I[\operatorname{sgn}(\hat a_m)
               =\operatorname{sgn}(a_m^*)],\\
q_m&=\left(1+
\frac{|\hat a_m-a_m^*|}{\max(|a_m^*|,\epsilon_a)}
\right)^{-1}.
\end{aligned}
\end{equation}
where $\epsilon_a=0.1$\,m for translation and $5^\circ$ for rotation. The step score is
\begin{equation}
g_m=\frac{1}{2}d_m(1+q_m).
\end{equation}
Wrong-direction estimates receive zero; correct-direction estimates receive between $0.5$ and $1$ according to magnitude. The benchmark action label selects the scored component for every step, and all 2{,}360 model-evaluation steps have a defined CMG score. The separately generated score-calibration references in Section~\ref{app:references} use a model-independent 2{,}358-step mask because two near-zero Chain actions are undefined in that evaluation run.

\paragraph{Temporal and protocol aggregation.}
For a case with $T_c$ steps, the case-level score $\mathrm{CMG}_c$ is the plain mean of its step scores,
\begin{equation}
\mathrm{CMG}_c=\frac{1}{T_c}\sum_{m=1}^{T_c}g_m.
\end{equation}
An atomic case reduces to its single step ($T_c=1$), while a three-step chain and a two-step cycle average their steps. Unlike the SSP pillars below, CMG uses no worst-step term: every step contributes equally within the case, and a within-case failure is diluted rather than pinned. Cases are then averaged within Atomic, Chain, and Cycle, and the three protocol means receive equal final weight. Thus a three-step case does not receive three times the weight of an atomic case.

\subsection{Scene State Preservation (SSP)}
\label{app:ssp_details}

\begin{figure*}[t]
\centering
\includegraphics[width=\textwidth]{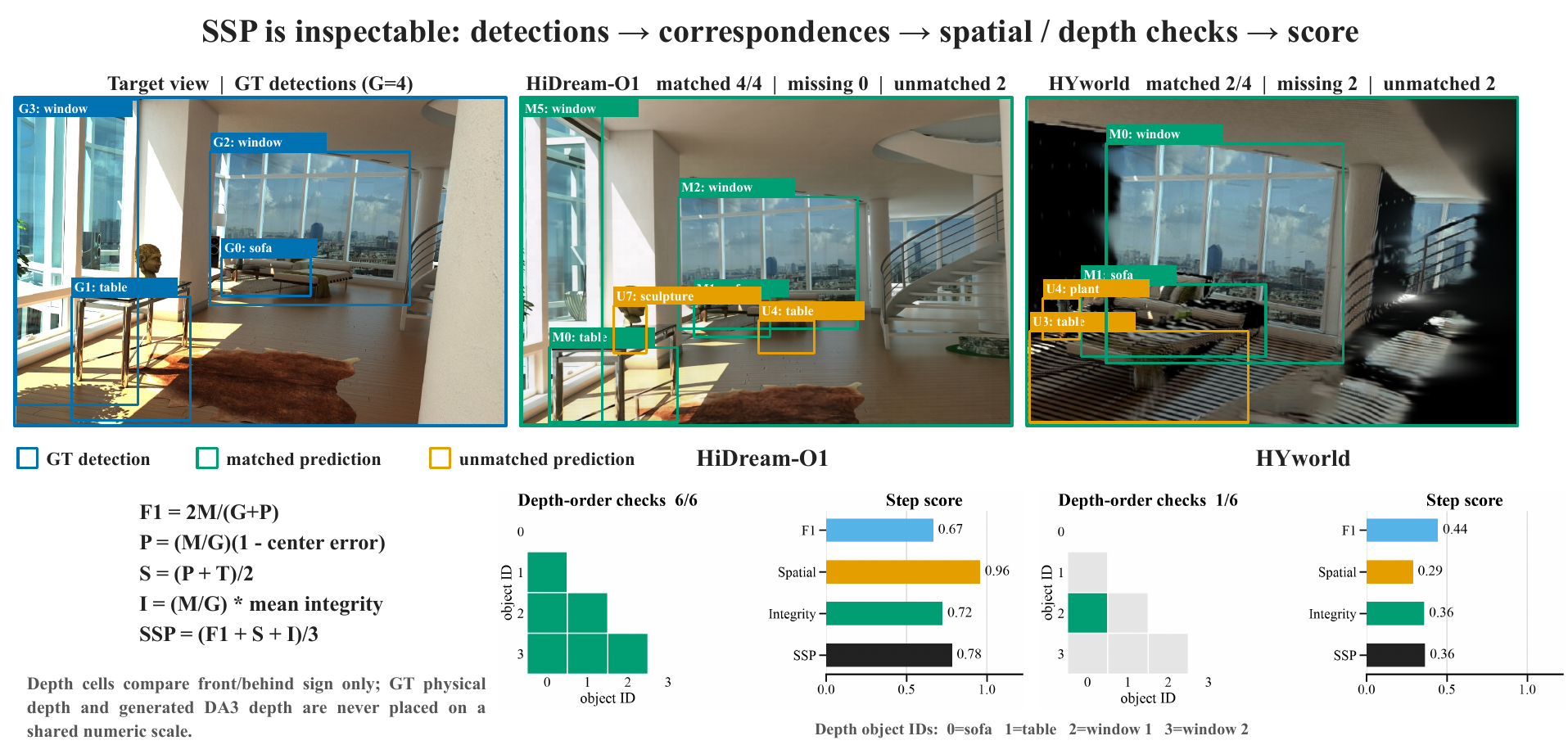}
\caption{Worked SSP example on one target step. \textbf{(a)}~GT detections define the evaluable object set; \textbf{(b,c)}~each model's generated view is scored by its matched, missing, and unmatched-prediction counts. \textbf{(d)}~The pillar formulas. \textbf{(e,f)}~The lower-triangular matrix records the front/behind check for every eligible GT pair (green correct, vermillion wrong or tied, light gray an eligible pair with a missing endpoint), and the bars give the resulting $F_1$, Spatial, Integrity, and SSP values. Depth cells compare front/behind sign only; GT physical depth and generated monocular depth are never placed on a shared numeric scale.}
\label{fig:ssp_walkthrough}
\end{figure*}

Figure~\ref{fig:ssp_walkthrough} traces the complete SSP computation on a single target step for two models: detector boxes on the target and generated views, the resulting one-to-one correspondences, the pairwise depth-order checks, and the three pillar scores that combine into the step score. The definitions below make each stage precise.

\paragraph{Detection and observable GT objects.}
Grounding DINO~\citep{liu2024grounding} detects objects in GT and generated images with the same prompt union. The union contains a fixed 36-class indoor vocabulary plus at most 20 normalized labels produced once for the GT target by Qwen3-VL~\citep{bai2025qwen3vltechnicalreport}; the deduplicated list is capped at 64 prompts and then held fixed for both images. Each evaluated step contributes 1--8 target-derived labels, producing 36--41 prompts, so neither cap is active. Box and text thresholds are $0.25/0.25$, and SAM3 is disabled; SSP therefore uses rectangular detector boxes rather than segmentation silhouettes.

GT poses and physical depth identify target objects that are supported by the supplied input views and pass the visibility and matchability filters. Let $G$ and $P$ be the resulting evaluable GT-object count and filtered generated-object count. A propose--demote--recover procedure based on Qwen3-VL, with DINOv3~\citep{simeoni2025dinov3} appearance verification, produces $M$ one-to-one matches. Unmatched generated proposals pass correspondence exclusion and box deduplication before being included in the false-positive count. They do not incur a second penalty because they already reduce precision through $P$ in $F_1$. Objects outside the shared prompt union are outside the declared open-vocabulary scope. A model-independent GT-only mask removes an entire case if any required step has $G=0$. The same mask excludes 44 cases and retains 1{,}356 cases/2{,}265 steps for all 18 evaluated systems.

\paragraph{Presence, position, and integrity.}
Object presence is
\begin{equation}
F_1=\frac{2M}{G+P}.
\end{equation}
For the matched set $\mathcal M$, normalized center error and coverage-aware position are
\begin{equation}
\begin{aligned}
\mathrm{PE}&=\frac{1}{M}\sum_{(i,j)\in\mathcal M}
\frac{\lVert\mathbf c_i^*-\hat{\mathbf c}_j\rVert_2}{\sqrt{H^2+W^2}},\\
\mathcal P&=\frac{M}{G}\operatorname{clip}(1-\mathrm{PE},0,1).
\end{aligned}
\end{equation}
The matched-object appearance score $\bar I$ combines DINO identity, structural SSIM, edge IoU, sharpness, color, and shape with weights $(0.30,0.20,0.15,0.12,0.13,0.10)$. If a signal is unavailable, the available weights are renormalized to sum to one. Coverage-aware integrity is
\begin{equation}
\mathcal I=\frac{M}{G}\bar I.
\end{equation}
If $M=0$, both $\mathcal P$ and $\mathcal I$ are zero. The factor $M/G$ ensures that unmatched GT objects receive no position or integrity credit.

\paragraph{Planar and depth topology.}
For GT object pair $(a,b)$, the planar score is the non-negative cosine agreement between GT and generated box-center offsets:
\begin{equation}
O^{xy}_{ab}=
\begin{cases}
\max\!\left(0,\cos(\mathbf c_b^*-\mathbf c_a^*,
\hat{\mathbf c}_b-\hat{\mathbf c}_a)\right), & a,b\text{ matched},\\
0, & \text{otherwise}.
\end{cases}
\end{equation}
The cosine uses the product of the two offset norms in its denominator. Benchmark validation rejects a GT offset norm below $10^{-8}$; a generated offset norm below $10^{-8}$ receives zero planar credit rather than an undefined cosine.
GT depth is the median physical RGB-D depth in the central 80\% of each frozen GT box. Generated depth is the median DA3 single-image depth in the central 80\% of each matched generated box. Both require at least 16 finite pixels in the numeric range $[0.05,20]$. For depths $z_a,z_b$, define
\begin{equation}
r_{ab}=\frac{z_a-z_b}{0.5(|z_a|+|z_b|)+\varepsilon}.
\end{equation}
A GT pair is depth-eligible when $|r_{ab}^{\mathrm{GT}}|>0.03$. Its depth score $O^z_{ab}\in\{0,1\}$ is one only if both endpoints are matched, generated depths are valid and non-tied under the same $0.03$ gate, and the front--behind sign agrees; otherwise it is zero. Only relative order is compared, so GT and generated depth need not share a metric scale.

The per-pair topology score is
\begin{equation}
T_{ab}=
\begin{cases}
\tfrac{1}{2}(O^{xy}_{ab}+O^z_{ab}), & \text{GT depth-eligible},\\
O^{xy}_{ab}, & \text{otherwise}.
\end{cases}
\end{equation}
For $G\ge2$,
\begin{equation}
\mathcal T=\binom{G}{2}^{-1}\sum_{1\le a<b\le G}T_{ab},
\qquad
\mathcal S=\frac{1}{2}(\mathcal P+\mathcal T).
\end{equation}
For $G=1$, $\mathcal S=\mathcal P$. An unmatched endpoint therefore receives zero in every incident topology pair.

\paragraph{Temporal and protocol aggregation.}
For a case with $T_c$ steps, SSP applies the mean-plus-worst operator separately to each pillar,
\begin{equation}
\operatorname{MPW}_c(x)=
\frac{1}{2}\left(\frac{1}{T_c}\sum_{m=1}^{T_c}x_m+\min_m x_m\right),
\end{equation}
and defines
\begin{equation}
\mathrm{SSP}_c=\frac{1}{3}\sum_{x\in\{F_1,\mathcal S,\mathcal I\}}
\operatorname{MPW}_c(x).
\end{equation}
Cases are averaged within protocol, and Atomic/Chain/Cycle receive equal final weight. MPW makes transient state loss visible without allowing longer cases to dominate by step count.

\subsection{Overall and auxiliary quality}

Let $\mathcal C_p$ be all cases in protocol $p$ and $\mathcal V_p$ its fixed SSP-valid subset. The reported scores are
\begin{equation}
\begin{aligned}
\mathrm{CMG}
&=\frac{1}{3}\sum_p\frac{1}{|\mathcal C_p|}
\sum_{c\in\mathcal C_p}\mathrm{CMG}_c,\\
\mathrm{SSP}
&=\frac{1}{3}\sum_p\frac{1}{|\mathcal V_p|}
\sum_{c\in\mathcal V_p}\mathrm{SSP}_c.
\end{aligned}
\end{equation}
\begin{equation}
\mathrm{Overall}=\frac{1}{2}(\mathrm{CMG}+\mathrm{SSP}).
\end{equation}
PSNR, LPIPS, NIQE~\citep{mittal2012making}, MUSIQ~\citep{ke2021musiq}, and CLIP-IQA~\citep{wang2022exploring} are auxiliary. RefSim averages min--max-normalized PSNR and inverted LPIPS; VisQual averages normalized inverted NIQE, MUSIQ, and CLIP-IQA. Both are oriented so higher is better and are excluded from Overall and model ranking.

\subsection{Full metric decomposition}

Tables~\ref{tab:cmg_decomposition} and~\ref{tab:ssp_decomposition} report the per-protocol submetrics for all 18 systems, exposing the structure hidden by the two headline averages. Two patterns hold throughout. First, direction accuracy is consistently higher than direction-gated magnitude, so the gap between the \emph{Dir.}\ and \emph{Gated} columns measures how much CMG credit comes from sense alone: GPT-Image-2 reaches 0.887 Atomic direction accuracy but only 0.582 gated magnitude, and every pose-free system shows the same ordering. Second, SSP degradation on Chain cases spreads across presence ($F_1$), spatial organization, and integrity rather than concentrating in one pillar, which is why no single submetric explains the Chain drop.

\begin{table*}[t]
\centering
\small
\setlength{\tabcolsep}{2.3pt}
\begin{tabular}{lrrrrrrrrr}
\toprule
\multirow{2}{*}{\textbf{Model}} &
\multicolumn{3}{c}{\textbf{Atomic}} &
\multicolumn{3}{c}{\textbf{Chain}} &
\multicolumn{3}{c}{\textbf{Cycle}} \\
\cmidrule(lr){2-4}\cmidrule(lr){5-7}\cmidrule(lr){8-10}
& CMG & Dir. & Gated & CMG & Dir. & Gated & CMG & Dir. & Gated \\
\midrule
GPT-Image-2 & 0.735 & 0.887 & 0.582 & 0.717 & 0.870 & 0.564 & 0.712 & 0.848 & 0.576 \\
Seedream-5 & 0.716 & 0.859 & 0.573 & 0.674 & 0.819 & 0.530 & 0.675 & 0.808 & 0.541 \\
Gemini-3-Pro & 0.598 & 0.723 & 0.473 & 0.603 & 0.753 & 0.454 & 0.578 & 0.700 & 0.456 \\
HiDream-O1 & 0.566 & 0.691 & 0.441 & 0.443 & 0.561 & 0.326 & 0.443 & 0.552 & 0.334 \\
FLUX.2-dev & 0.560 & 0.680 & 0.440 & 0.362 & 0.467 & 0.257 & 0.421 & 0.525 & 0.317 \\
HunyuanImage-3.0 & 0.524 & 0.647 & 0.401 & 0.413 & 0.529 & 0.296 & 0.450 & 0.565 & 0.335 \\
Qwen-Image-Edit & 0.568 & 0.685 & 0.450 & 0.409 & 0.531 & 0.288 & 0.432 & 0.540 & 0.325 \\
Step1X-Edit & 0.317 & 0.421 & 0.214 & 0.306 & 0.406 & 0.205 & 0.271 & 0.360 & 0.182 \\
BAGEL-7B-MoT & 0.460 & 0.566 & 0.355 & 0.431 & 0.539 & 0.323 & 0.414 & 0.515 & 0.314 \\
OmniGen2 & 0.495 & 0.616 & 0.374 & 0.392 & 0.502 & 0.282 & 0.382 & 0.483 & 0.280 \\
ACE++ & 0.409 & 0.521 & 0.297 & 0.384 & 0.499 & 0.269 & 0.382 & 0.498 & 0.267 \\
FireRed-Edit & 0.480 & 0.603 & 0.357 & 0.383 & 0.496 & 0.269 & 0.356 & 0.456 & 0.256 \\
ICEdit & 0.359 & 0.469 & 0.249 & 0.330 & 0.440 & 0.219 & 0.325 & 0.423 & 0.227 \\
Emu3.5 & 0.415 & 0.499 & 0.330 & 0.469 & 0.573 & 0.365 & 0.446 & 0.531 & 0.360 \\
Kling-v2.1-Autoreg & 0.746 & 0.956 & 0.536 & 0.724 & 0.955 & 0.494 & 0.760 & 0.965 & 0.556 \\
Seedance-1.0-Pro-Fast-Autoreg & 0.609 & 0.756 & 0.462 & 0.568 & 0.738 & 0.397 & 0.645 & 0.792 & 0.498 \\
\textit{HY-WorldMirror-2.0} & 0.839 & 0.989 & 0.688 & 0.883 & 0.981 & 0.784 & 0.821 & 0.983 & 0.658 \\
\textit{Lingbot-World} & 0.765 & 0.915 & 0.616 & 0.742 & 0.877 & 0.606 & 0.735 & 0.892 & 0.578 \\
\bottomrule
\end{tabular}
\caption{CMG decomposition for all 18 systems. Gated is direction-gated magnitude $dq$; CMG is $(d+dq)/2$. Values use the official case-first protocol aggregation and match the headline CMG scores in the main text. Italicized rows are pose-conditioned references reported separately from the primary pose-free ranking.}
\label{tab:cmg_decomposition}
\end{table*}

\begin{table*}[t]
\centering
\small
\setlength{\tabcolsep}{1.75pt}
\begin{tabular}{lrrrrrrrrrrrr}
\toprule
\multirow{2}{*}{\textbf{Model}} &
\multicolumn{4}{c}{\textbf{Atomic}} &
\multicolumn{4}{c}{\textbf{Chain}} &
\multicolumn{4}{c}{\textbf{Cycle}} \\
\cmidrule(lr){2-5}\cmidrule(lr){6-9}\cmidrule(lr){10-13}
& SSP & F1 & Sp. & Int. & SSP & F1 & Sp. & Int. & SSP & F1 & Sp. & Int. \\
\midrule
GPT-Image-2 & 0.66 & 0.72 & 0.66 & 0.59 & 0.55 & 0.61 & 0.54 & 0.50 & 0.60 & 0.65 & 0.60 & 0.55 \\
Seedream-5 & 0.61 & 0.70 & 0.60 & 0.54 & 0.48 & 0.57 & 0.45 & 0.42 & 0.53 & 0.61 & 0.51 & 0.47 \\
Gemini-3-Pro & 0.62 & 0.70 & 0.60 & 0.55 & 0.49 & 0.56 & 0.45 & 0.45 & 0.55 & 0.62 & 0.53 & 0.50 \\
HiDream-O1 & 0.61 & 0.69 & 0.61 & 0.54 & 0.50 & 0.57 & 0.48 & 0.45 & 0.57 & 0.63 & 0.57 & 0.52 \\
FLUX.2-dev & 0.61 & 0.69 & 0.60 & 0.54 & 0.49 & 0.56 & 0.47 & 0.45 & 0.56 & 0.62 & 0.55 & 0.51 \\
HunyuanImage-3.0 & 0.62 & 0.69 & 0.61 & 0.55 & 0.41 & 0.49 & 0.39 & 0.36 & 0.57 & 0.64 & 0.56 & 0.51 \\
Qwen-Image-Edit & 0.62 & 0.69 & 0.61 & 0.56 & 0.40 & 0.48 & 0.37 & 0.35 & 0.54 & 0.61 & 0.53 & 0.49 \\
Step1X-Edit & 0.60 & 0.69 & 0.58 & 0.54 & 0.49 & 0.57 & 0.45 & 0.45 & 0.63 & 0.70 & 0.61 & 0.58 \\
BAGEL-7B-MoT & 0.55 & 0.64 & 0.52 & 0.49 & 0.39 & 0.47 & 0.34 & 0.35 & 0.46 & 0.54 & 0.42 & 0.41 \\
OmniGen2 & 0.52 & 0.61 & 0.49 & 0.47 & 0.35 & 0.43 & 0.31 & 0.31 & 0.49 & 0.56 & 0.46 & 0.44 \\
ACE++ & 0.44 & 0.55 & 0.40 & 0.38 & 0.30 & 0.38 & 0.26 & 0.26 & 0.48 & 0.57 & 0.44 & 0.42 \\
FireRed-Edit & 0.50 & 0.60 & 0.46 & 0.43 & 0.21 & 0.27 & 0.18 & 0.17 & 0.42 & 0.51 & 0.39 & 0.36 \\
ICEdit & 0.33 & 0.43 & 0.30 & 0.28 & 0.24 & 0.30 & 0.20 & 0.20 & 0.31 & 0.39 & 0.29 & 0.27 \\
Emu3.5 & 0.32 & 0.44 & 0.25 & 0.28 & 0.10 & 0.15 & 0.07 & 0.09 & 0.19 & 0.26 & 0.14 & 0.16 \\
Kling-v2.1-Autoreg & 0.56 & 0.64 & 0.54 & 0.51 & 0.32 & 0.40 & 0.28 & 0.29 & 0.51 & 0.57 & 0.49 & 0.47 \\
Seedance-1.0-Pro-Fast-Autoreg & 0.54 & 0.62 & 0.52 & 0.48 & 0.36 & 0.44 & 0.32 & 0.32 & 0.44 & 0.51 & 0.42 & 0.40 \\
\textit{HY-WorldMirror-2.0} & 0.48 & 0.58 & 0.44 & 0.43 & 0.42 & 0.50 & 0.38 & 0.37 & 0.55 & 0.61 & 0.52 & 0.51 \\
\textit{Lingbot-World} & 0.60 & 0.68 & 0.58 & 0.54 & 0.44 & 0.52 & 0.41 & 0.40 & 0.55 & 0.62 & 0.53 & 0.50 \\
\bottomrule
\end{tabular}
\caption{SSP decomposition on the fixed GT-only mask for all 18 systems. Spatial (Sp.) and Integrity (Int.) are coverage-aware All-GT quantities, so unmatched GT objects contribute zero. Italicized rows are pose-conditioned references.}
\label{tab:ssp_decomposition}
\end{table*}

The pose-conditioned references confirm the interpretation. Privileged 6-DoF control lifts HY-WorldMirror-2.0 and Lingbot-World well above the pose-free field on CMG, yet their SSP remains within the pose-free range because state preservation depends on target-view content that trajectory control alone does not guarantee.

\section{Human Validation and Ranking Uncertainty}
\label{app:human}

\subsection{Blinded protocol}

The primary study samples 180 protocol-stratified parent cases (80 Atomic, 60 Chain, 40 Cycle), yielding 340 steps. Three rubric-trained annotators independently rank the same six anonymized systems per step, with ties allowed. CMG judgments emphasize action direction, magnitude, and target-view agreement; SSP judgments emphasize target-visible objects, image-plane organization, relative depth, and recognizable appearance. Visual polish is not a primary criterion, and model identities and automatic scores are hidden.

Ties receive average ranks and rank $r$ maps to utility $(6-r)/5$. Utility is first averaged across annotators for each item--model pair. Automatic metrics are recomputed on identical examples using the official step-to-case-to-protocol aggregation, SSP mask, and mean-plus-worst rule. The matched data contain exactly $340\times6=2{,}040$ model--step rows. Three additional annotators independently re-evaluate a stratified 102-step subset (24 Atomic, 54 Chain, 24 Cycle).

\subsection{Agreement at system and item level}

\begin{table*}[t]
\centering
\small
\setlength{\tabcolsep}{4pt}
\resizebox{\textwidth}{!}{%
\begin{tabular}{llll}
\toprule
\textbf{Evidence} & \textbf{Axis} & \textbf{Statistic} & \textbf{Scope} \\
\midrule
Aggregate rank & CMG & $\rho=0.943,\ \tau_b=0.867$, exact $p=0.0167$ & 6 systems, 340 steps \\
Aggregate rank & SSP & $\rho=0.943,\ \tau_b=0.867$, exact $p=0.0167$ & 6 systems, 340 steps \\
Aggregate rank & Overall & $\rho=0.943,\ \tau_b=0.867$, exact $p=0.0167$ & 6 systems, 340 steps \\
Item alignment & CMG & mean $\rho=0.348$ [0.302,0.392]; pairwise 0.639 [0.621,0.657] & 340 steps/180 cases \\
Item alignment & SSP & mean $\rho=0.334$ [0.289,0.377]; pairwise 0.634 [0.615,0.653] & $\rho$: 333/177; pairwise: 340/180 \\
Reannotation & CMG & ICC(A,1)=0.596; ICC(A,3)=0.816; cross-group $\rho=1.000$ & 102 steps, 6 systems \\
Reannotation & SSP & ICC(A,1)=0.632; ICC(A,3)=0.838; cross-group $\rho=0.943$ & 102 steps, 6 systems \\
\bottomrule
\end{tabular}
}
\caption{Human-validation evidence. Brackets are 95\% parent-case cluster-bootstrap intervals from 10{,}000 resamples (seed 20260619). Pairwise agreement gives half credit to automatic ties on human-strict pairs.}
\label{tab:human_evidence}
\end{table*}

The automatic ordering is GPT-Image-2, Seedream-5, Gemini-3-Pro, Qwen-Image-Edit, BAGEL-7B-MoT, and OmniGen2; human consensus swaps only the last two. The high six-system correlation supports aggregate ordering on this matched panel. The moderate item-level values---including the reported 0.348 CMG mean correlation---show why the metrics are used for case-, protocol-, and system-level comparison rather than exact per-example grading or separate validation of every SSP subcomponent.

\subsection{Comparison with conventional image metrics}
\label{app:human_metric_baselines}

To test whether the aggregate human agreement merely reflects generic image quality or reference similarity, we recompute each candidate metric on the identical six-system, 340-step panel and apply the same step-to-case-to-protocol aggregation. Table~\ref{tab:human_metric_baselines} orients NIQE and LPIPS so that higher values are better. Overall best matches the human system ordering, followed by CLIP-IQA. With only six systems, these differences are descriptive rather than a powered significance claim.

\begin{table}[t]
\centering
\small
\setlength{\tabcolsep}{5pt}
\begin{tabular}{lrr}
\toprule
\textbf{Automatic metric} & \textbf{Spearman $\rho$} & \textbf{Kendall $\tau_b$} \\
\midrule
\textbf{Overall} & \textbf{0.943} & \textbf{0.867} \\
CLIP-IQA & 0.886 & 0.733 \\
MUSIQ & 0.714 & 0.467 \\
PSNR & 0.600 & 0.333 \\
$-$NIQE & 0.429 & 0.467 \\
$-$LPIPS & 0.314 & 0.200 \\
\bottomrule
\end{tabular}
\caption{System-level agreement with aggregate human ordering on the same six-system panel. Negative NIQE/LPIPS orient all metrics as higher-is-better. The comparison is descriptive because $n=6$.}
\label{tab:human_metric_baselines}
\end{table}

The component-level boundary is equally important. At item level, CMG reaches mean $\rho=0.348$, while negative dominant-action absolute error reaches 0.371; their paired parent-case-bootstrap difference is $[-0.039,-0.007]$. SSP reaches 0.334, comparable to Spatial (0.357), Integrity (0.356), and PSNR (0.356), whose paired differences from SSP include zero. The composite scores are therefore justified as pre-defined summaries for aggregate capability comparison, not as universally superior per-item perceptual metrics.

\subsection{Paired case-bootstrap uncertainty}
\label{app:uncertainty}

We resample parent cases within subtype, use identical draws across systems, and recompute the complete step-to-case-to-protocol aggregation. CMG and SSP use independent subtype-stratified draws; Overall averages their replicate scores. Table~\ref{tab:leaderboard_ci} reports 2{,}000 percentile-bootstrap resamples (seed 20260619). The one-sided $p_{\mathrm{flip}}$ is the fraction of paired replicates in which each displayed row reverses or ties with the following row.

\begin{table*}[t]
\centering
\small
\setlength{\tabcolsep}{3pt}
\begin{tabular}{lrrrr}
\toprule
\textbf{System} & \textbf{Overall [95\% CI]} & \textbf{CMG [95\% CI]} & \textbf{SSP [95\% CI]} & \textbf{$p_{\mathrm{flip}}$} \\
\midrule
\textit{HY-WorldMirror-2.0} & 0.664 [0.655,0.673] & [0.838,0.857] & [0.466,0.495] & 0.344 \\
GPT-Image-2 & 0.662 [0.654,0.670] & [0.708,0.734] & [0.592,0.612] & 0.001 \\
\textit{Lingbot-World} & 0.639 [0.629,0.648] & [0.734,0.760] & [0.517,0.544] & 0.001 \\
Seedream-5 & 0.615 [0.606,0.625] & [0.673,0.703] & [0.530,0.554] & 0.028 \\
Kling-v2.1-Autoreg & 0.605 [0.597,0.612] & [0.735,0.753] & [0.453,0.477] & 0.000 \\
Gemini-3-Pro & 0.572 [0.561,0.583] & [0.576,0.610] & [0.538,0.563] & 0.000 \\
Seedance-1.0-Pro-Fast-Autoreg & 0.528 [0.517,0.538] & [0.591,0.623] & [0.435,0.461] & 0.290 \\
HiDream-O1 & 0.524 [0.513,0.534] & [0.468,0.501] & [0.552,0.575] & 0.000 \\
FLUX.2-dev & 0.501 [0.491,0.511] & [0.431,0.464] & [0.543,0.565] & 0.325 \\
HunyuanImage-3.0 & 0.498 [0.487,0.509] & [0.445,0.479] & [0.521,0.545] & 0.314 \\
Qwen-Image-Edit & 0.495 [0.485,0.505] & [0.452,0.487] & [0.508,0.531] & 0.000 \\
BAGEL-7B-MoT & 0.451 [0.440,0.461] & [0.417,0.453] & [0.452,0.479] & 0.035 \\
OmniGen2 & 0.438 [0.427,0.449] & [0.406,0.439] & [0.440,0.465] & 0.421 \\
Step1X-Edit & 0.436 [0.426,0.446] & [0.283,0.314] & [0.563,0.586] & 0.000 \\
ACE++ & 0.400 [0.389,0.410] & [0.376,0.408] & [0.394,0.422] & 0.080 \\
FireRed-Edit & 0.390 [0.380,0.399] & [0.391,0.422] & [0.361,0.385] & 0.000 \\
Emu3.5 & 0.324 [0.314,0.335] & [0.425,0.459] & [0.196,0.215] & 0.157 \\
ICEdit & 0.316 [0.306,0.327] & [0.321,0.354] & [0.282,0.309] & -- \\
\bottomrule
\end{tabular}
\caption{Case-bootstrap uncertainty. Italicized rows are pose-conditioned references, not members of the primary pose-free ranking. Adjacent rows with $p_{\mathrm{flip}}\ge0.05$ should be read as local ties.}
\label{tab:leaderboard_ci}
\end{table*}

The median Overall interval half-width is approximately 0.010. In particular, HY-WorldMirror-2.0 and GPT-Image-2 are not statistically ordered ($p_{\mathrm{flip}}=0.344$), nor are several close middle and lower pairs. These intervals quantify benchmark case sampling only: one frozen output exists per model--step, so they do not include stochastic generation variance.

\section{Evaluator Robustness}
\label{app:robustness}

Each intervention changes one evaluator component while holding generated outputs, GT actions, and remaining aggregation fixed. The tests establish broad rank robustness, not numerical equivalence or ground-truth accuracy of one learned evaluator.

\paragraph{CMG pose and preprocessing.}
Replacing DA3 with VGGT-1B on a controlled six-model panel yields $\rho=0.943$ and $\tau_b=0.867$, with unchanged Top-1 and Top-3. Native-aspect versus GT-aspect center-crop preprocessing changes CMG by 0.0019 on average and 0.0044 at most, with $\rho=\tau_b=1.000$. Across the 16 pose-free systems, direction-only aggregation agrees with full CMG at $\rho=0.997$ and $\tau_b=0.983$. The broad ordering is therefore not created by the magnitude term alone.

\paragraph{SSP geometry and detection.}
Replacing generated-object DA3 depth with VGGT on an eight-model, 342-case common panel preserves the complete ordering ($\rho=1.000$) while shifting mean macro SSP by $-0.0124$. Jointly changing detector box/text thresholds from $0.25/0.25$ to $0.20/0.20$ and $0.30/0.30$ gives Overall $\rho=0.929/0.976$, unchanged Top-1, and maximum absolute score change 0.00873. Comparing the 36 fixed prompts with their deterministic union with frozen GT labels gives $\rho=0.976$, unchanged Top-1, and maximum change 0.00542. Common masks prevent alternatives from gaining evaluation coverage.

\paragraph{Correspondence and temporal aggregation.}
Replacing propose--demote--recover with DINO-Hungarian assignment preserves Overall Top-1 but yields $\rho=0.857$ overall and $0.595$ on Cycle; assignments and close ranks remain sensitive. Without human pair labels, this does not establish either matcher as more accurate. Replacing pillar-wise MPW with mean-only aggregation gives SSP $\rho=0.994$ and Overall $\rho=0.996$ on 18 systems with unchanged Top-3.

\section{Benchmark-Design and Sampling Robustness}
\label{app:benchmark_robustness}

Evaluator ablations ask whether a learned backend creates the ranking. This section asks a complementary benchmark-design question: whether the principal conclusions depend on the temporal functional, case-independence assumption, dominant data source, score weights, or selected benchmark size. All analyses reuse the frozen outputs and the 16 pose-free primary systems.

\subsection{Matched temporal aggregation}
\label{app:temporal_aggregation}

The official axes intentionally answer different questions: CMG averages action execution across steps, whereas SSP uses $\operatorname{MPW}(x)=\tfrac12(\operatorname{mean}(x)+\min(x))$ within each pillar to expose a corrupted intermediate view. Comparing their Atomic-to-Chain drops therefore mixes capability and functional sensitivity. We repeat the comparison with both axes using the same functional. Confidence intervals use 5{,}000 dataset-stratified scene-cluster bootstrap resamples shared across systems and axes.

\begin{table*}[t]
\centering
\small
\setlength{\tabcolsep}{4pt}
\begin{tabular}{lrrrrr}
\toprule
\textbf{CMG/SSP temporal rule} & \textbf{CMG drop} & \textbf{SSP drop} & \textbf{SSP$-$CMG} & \textbf{Gap 95\% scene CI} & \textbf{SSP larger} \\
\midrule
Mean / MPW (official) & 0.0593 & 0.1643 & $+0.1050$ & [0.0837, 0.1250] & 14/16 \\
Mean / Mean & 0.0593 & 0.0792 & $+0.0198$ & [$-0.0007$, 0.0396] & 11/16 \\
MPW / MPW & 0.1826 & 0.1643 & $-0.0182$ & [$-0.0396$, 0.0021] & 5/16 \\
\bottomrule
\end{tabular}
\caption{Model-mean Atomic-to-Chain drops under matched and official temporal aggregation. ``SSP larger'' counts systems whose SSP drop exceeds their CMG drop.}
\label{tab:matched_temporal}
\end{table*}

The official $0.059/0.164$ drops are correct descriptive statistics, but the $0.105$ contrast is not aggregation-invariant. Under matched means, SSP still has the larger point drop by 0.020, but its interval slightly crosses zero; under matched MPW, CMG has the larger point drop by 0.018. A shared temporal-weight sweep changes the sign near a minimum-step weight of 0.26. The robust finding is that Chain degrades both axes. The additional official SSP drop quantifies the tail failure that its worst-step-sensitive definition is designed to expose; it should not be interpreted alone as evidence for an independent state-memory mechanism.

\subsection{Scene-cluster ranking uncertainty}
\label{app:scene_cluster}

The primary case bootstrap stratifies by nine instruction subtypes. Because 1{,}400 cases come from 771 scenes, we additionally resample scenes within each source dataset and retain every case from a selected scene; 5{,}000 shared resamples preserve paired model comparisons. For the 16-system primary ranking, the median Overall interval half-width increases from 0.0100 under case sampling to 0.0117 under scene sampling, a median factor of 1.180. The pose-free Top-1 comparison remains unambiguous ($p_{\mathrm{flip}}=0$), and the Seedream--Kling Top-3 boundary has $p_{\mathrm{flip}}=0.035$. However, six of 15 adjacent pairs have $p_{\mathrm{flip}}\ge0.05$. These results support the leading system and broad tiers while reinforcing that several middle and lower neighbors are local ties.

\subsection{Source coverage and leave-one-source-out stability}
\label{app:source_robustness}

Source coverage is feasibility-constrained rather than balanced: ScanNet++ contributes 1{,}003 of 1{,}400 cases, while Matterport3D contributes 22. Hypersim and Matterport3D contain no Chain cases, so we compare sources on their common Atomic+Cycle protocols instead of conflating source and protocol composition.

All four source-specific panels retain GPT-Image-2 as Top-1, with common-protocol rank correlations of $\rho=0.897$--$0.959$ against the full-data Atomic+Cycle ordering. Removing any one source leaves $\rho=0.968$--$1.000$ against the full primary ranking; even after removing ScanNet++, which contributes 71.6\% of the cases, the pose-free Top-3 remains GPT-Image-2, Seedream-5, and Kling, and equal weighting of the four sources also retains GPT-Image-2 as Top-1. The broad ordering is therefore not solely a ScanNet++ artifact.

\subsection{Score-weight sensitivity and benchmark size}
\label{app:weight_sample_robustness}

Let $\operatorname{Overall}_{\lambda}=\lambda\operatorname{CMG}+(1-\lambda)\operatorname{SSP}$. For $\lambda\in[0.25,0.75]$, the minimum rank correlation with equal weighting is $\rho=0.926$ and GPT-Image-2 remains Top-1; it remains Top-1 over the wider $[0,0.85]$ range, while Kling becomes Top-1 at the motion-dominant $\lambda=0.90$. Equal weighting is therefore not uniquely necessary, but extreme preferences appropriately change the ranking. For SSP, 20{,}000 $\operatorname{Dirichlet}(1,1,1)$ pillar-weight draws retain GPT-Image-2 as Top-1 in 100\% of draws, and the 95\% range of rank correlation with equal-pillar SSP is $[0.991,1.000]$. Dropping any one pillar gives $\rho=0.997$, 0.997, and 1.000, with unchanged Top-1.

Subsampling within the nine instruction subtypes (1{,}000 no-replacement draws per size) shows the 16-system ordering is already stable at 140 cases (median $\rho=0.985$, Top-1 match 1.000) and that the full 1{,}400 cases mainly reduce score error (MAE $0.0121\to0.0023$). This post-hoc check is not a power analysis for future systems and does not guarantee every adjacent order.

\section{Claim-Focused Diagnostics}
\label{app:diagnostics}

All diagnostics reuse frozen evaluated outputs. They explain score behavior and do not create additional leaderboards or causal claims.

\subsection{Controlled Context-\texorpdfstring{$K$}{K}: fixed objects and view confusion}
\label{app:context_k}

Seven native multi-image models are evaluated on 70 matched base cases at $K=1$--4; a common mask retains 69 cases/116 steps. Under the model-balanced aggregation used in Figure~\ref{fig:mechanistic_summary}(b), CMG at $K=1,2,3,4$ is 0.519, 0.551, 0.580, and 0.582. Native SSP evaluates the object set supported at each $K$, which changes on 45/118 base steps as auxiliaries add evidence. We therefore recompute SSP on 440 exact GT label--box instances present at every $K$. ``Fixed/filtered'' also filters predictions to this intersection; ``fixed/all predictions'' retains all generated predictions and is the conservative variant because unmatched predictions outside the intersection still reduce precision.

\begin{table}[t]
\centering
\small
\setlength{\tabcolsep}{1.3pt}
\begin{tabular}{lrrrrr}
\toprule
$K$ & Native & Fixed/filtered & Fixed/all & Source recall & Aux.-only recall \\
\midrule
1 & 0.539 & 0.584 & 0.537 & 0.725 & -- \\
2 & 0.533 & 0.582 & 0.531 & 0.735 & 0.632 \\
3 & 0.514 & 0.556 & 0.506 & 0.725 & 0.650 \\
4 & 0.504 & 0.549 & 0.500 & 0.710 & 0.631 \\
\bottomrule
\end{tabular}
\caption{Model-balanced Context-$K$ results. Source recall uses the stable cross-$K$ intersection; auxiliary-only objects are $\mathcal O_K\setminus\mathcal O_1$.}
\label{tab:context_fixed}
\end{table}

From $K=1$ to $K=4$, native, fixed/filtered, and fixed/all-prediction SSP change by $-0.0354$, $-0.0347$, and $-0.0368$, respectively. The latter two changes have 95\% CIs of [$-0.0653,-0.0033$] and [$-0.0644,-0.0084$], with one-sided $p_{\ge0}=0.0147$ and $0.0047$. Fixed-object F1, Spatial, and Integrity change by $-0.0433$, $-0.0327$, and $-0.0280$. Intervals use 10{,}000 subtype-stratified parent-case bootstrap resamples shared across $K$ and models (seed 20260729). Thus the small decline persists under a fixed-object comparison, while stable source-visible recall changes only from 0.725 to 0.710 and auxiliary-only recall remains 0.631--0.650. The result indicates imperfect integration of added views, not that context is intrinsically harmful.

CLIP ViT-L/14@336 provides a complementary view-confusion diagnostic. At $K=4$, 62.2\% of step-1 outputs are feature-nearest to an auxiliary but only 0.6\% are near-pixel copies (RGB RMSE $\le0.02$ after deterministic resizing). Strict auxiliary-confused outputs score 0.064 lower SSP than other rows (95\% CI [0.009,0.119]); across paired $K=2$--4 rows, auxiliary attraction and SSP change correlate only weakly ($\rho=-0.109$). This is evidence of partial feature-space attraction, not widespread literal copying.

\subsection{Motion magnitude and residual pose}

Translation and rotation steps are binned by tertiles of absolute GT action magnitude. Under/near/over denote predicted-to-GT magnitude ratios below 0.8, within $[0.8,1.2]$, and above 1.2, conditional on correct direction.

\begin{table}[t]
\centering
\small
\setlength{\tabcolsep}{2.5pt}
\begin{tabular}{llrrrr}
\toprule
\textbf{Family} & \textbf{Bin} & \textbf{Direction} & \textbf{Under} & \textbf{Near} & \textbf{Over} \\
\midrule
Translation & Small & 0.565 & 0.611 & 0.083 & 0.306 \\
Translation & Medium & 0.609 & 0.689 & 0.093 & 0.218 \\
Translation & Large & 0.610 & 0.707 & 0.117 & 0.175 \\
Rotation & Small & 0.663 & 0.561 & 0.106 & 0.333 \\
Rotation & Medium & 0.655 & 0.595 & 0.137 & 0.268 \\
Rotation & Large & 0.659 & 0.680 & 0.133 & 0.187 \\
\bottomrule
\end{tabular}
\caption{Magnitude diagnostics over 16 pose-free systems. Parent cases are averaged before systems.}
\label{tab:magnitude_compact}
\end{table}

Small-rotation direction accuracy is 0.663 versus the 0.525 majority-sign baseline. For medium and large actions, 59.5--70.7\% of direction-correct outputs under-execute and only 9.3--13.7\% lie within 20\% of the requested magnitude. Full-pose records also show that dominant-axis success is not full-pose fidelity: GPT-Image-2 has mean translation-direction and SO(3) errors of $43.68^\circ$ and $15.80^\circ$; pose-conditioned HY-WorldMirror-2.0 reaches $10.51^\circ$ translation-direction and $2.52^\circ$ SO(3) error but a 1.491\,m full-translation error due to over-scaling.

\subsection{Cycle return behavior}
\label{app:cycle_behavior}

Cycle tests whether action responsiveness persists after one self-generated step. For this diagnostic only, we define the absolute response ratio $r=|\hat a_m|/|a_m^*|$ on the instructed dominant component and call a response near-static when $r\leq0.2$. This describes estimated camera response, not pixel identity.

\begin{table}[t]
\centering
\small
\setlength{\tabcolsep}{3.5pt}
\begin{tabular}{lrrrr}
\toprule
\textbf{Step} & \textbf{$N$} & \textbf{Median $r$} & \textbf{Near-static} & \textbf{Direction} \\
\midrule
Outbound & 3{,}840 & 0.412 & 37.4\% & 66.4\% \\
Return & 3{,}840 & 0.013 & 66.3\% & 53.1\% \\
\bottomrule
\end{tabular}
\caption{Observed Cycle action behavior across the 16 pose-free systems. $N$ counts transitions. Near-static is a descriptive $r\leq0.2$ diagnostic; Direction is signed dominant-action accuracy.}
\label{tab:cycle_return_behavior}
\end{table}

Ten of the 16 systems are near-static on more than half of their return steps. Among the 3{,}840 Cycle cases, 32.2\% are near-static on both outbound and return, and another 34.1\% move outbound but become near-static on return. The first group barely executes the requested camera cycle; the second moves to an intermediate viewpoint but then fails to invert the action. Individual outputs also include wrong-direction and over-motion failures, so the aggregate should not be read as uniform behavior across models. These observations motivate free-running inverse-action supervision; they do not establish whether the suppression arises from the generator, accumulated visual drift, or the learned pose evaluator.

\subsection{SSP conditioned on pose accuracy}

To separate wrong-view and state effects observationally, Table~\ref{tab:ssp_pose_strata} pools transitions from the 16 pose-free systems under increasingly strict pose criteria. Confidence intervals cluster 10{,}000 bootstrap resamples by parent case (seed 20260729).

\begin{table*}[t]
\centering
\small
\setlength{\tabcolsep}{4pt}
\begin{tabular}{lrrrrr}
\toprule
\textbf{Subset} & \textbf{$N$} & \textbf{SSP [95\% CI]} & \textbf{$F_1$} & \textbf{Spatial} & \textbf{Integrity} \\
\midrule
All valid & 36{,}240 & 0.510 [0.504,0.517] & 0.589 & 0.489 & 0.452 \\
CMG $\ge 0.8$ & 8{,}956 & 0.561 [0.553,0.569] & 0.634 & 0.551 & 0.499 \\
Loose full pose & 1{,}515 & 0.595 [0.579,0.610] & 0.669 & 0.589 & 0.526 \\
Strict full pose & 418 & 0.597 [0.571,0.624] & 0.668 & 0.591 & 0.533 \\
\bottomrule
\end{tabular}
\caption{Target-view SSP conditioned on pose accuracy. Loose thresholds are translation-direction/SO(3)/translation-norm errors $\le30^\circ/15^\circ/0.3$\,m; strict thresholds are $\le20^\circ/10^\circ/0.2$\,m.}
\label{tab:ssp_pose_strata}
\end{table*}

SSP rises with pose accuracy but remains near 0.60 in the strictest stratum, so wrong view is not the only observed source of SSP deficit.

\subsection{Object retention by projected size}

On 1{,}351 cases/2{,}255 steps with identical GT instance catalogs across systems, confirmed retention is 0.604 [0.592,0.615], 0.635 [0.623,0.645], and 0.696 [0.686,0.705] for small, medium, and large projected objects. Conditional position varies by at most 0.011 and integrity by 0.006, so the 0.093 large--small gap mainly reflects detection/matching-confirmed retention. From Chain step~1 to step~3, retention drops by 0.195 for large furniture and 0.191 for small/portable objects; cumulative degradation is not specific to small objects.

\subsection{Off-axis and full-pose error}
\label{app:full_pose}

CMG deliberately scores only the frozen dominant action component. The corrected all-step pose records also permit a stricter diagnostic of unintended or mismatched residual motion. For a translation instruction whose dominant axis is $k\in\{x,z\}$, let $\mathbf e_t=\hat{\mathbf t}-\mathbf t^*$ and define
\begin{equation}
E_t^{\mathrm{full}}=\lVert\mathbf e_t\rVert_2,\qquad
E_t^{\mathrm{off}}=
\left\lVert(e_{t,j})_{j\ne k}\right\rVert_2.
\end{equation}
This subtracts the complete GT translation before measuring the residual, so legitimate off-axis motion in the real trajectory is not itself penalized. We additionally report the angle between the full predicted and GT translation vectors. For rotation, $E_R^{\mathrm{full}}$ is the SO(3) geodesic error. The cross-axis diagnostic is $|\hat\theta-\theta^*|$ for a yaw instruction and $|\hat\psi-\psi^*|$ for a pitch instruction; the geodesic term additionally captures roll error and axis coupling.

\begin{table*}[t]
\centering
\small
\setlength{\tabcolsep}{3.4pt}
\begin{tabular}{lrrrrr}
\toprule
\textbf{Model} &
\textbf{Full trans. (m)} &
\textbf{Off-axis trans. (m)} &
\textbf{Trans. dir. ($^\circ$)} &
\textbf{SO(3) error ($^\circ$)} &
\textbf{Cross-axis rot. ($^\circ$)} \\
\midrule
GPT-Image-2 & 0.682 & 0.363 & 43.68 & 15.80 & 6.41 \\
Seedream-5 & 0.715 & 0.302 & 60.43 & 15.40 & 6.95 \\
Gemini-3-Pro & 1.004 & 0.466 & 69.02 & 24.28 & 9.93 \\
HiDream-O1 & 0.865 & 0.364 & 84.35 & 19.66 & 7.57 \\
FLUX.2-dev & 1.077 & 0.570 & 81.76 & 23.08 & 8.77 \\
HunyuanImage-3.0 & 0.784 & 0.272 & 80.03 & 18.86 & 6.40 \\
Qwen-Image-Edit & 0.903 & 0.411 & 80.00 & 23.57 & 7.77 \\
BAGEL-7B-MoT & 0.963 & 0.479 & 73.64 & 30.83 & 9.30 \\
OmniGen2 & 0.875 & 0.374 & 83.99 & 21.91 & 8.61 \\
Step1X-Edit & 0.784 & 0.221 & 90.62 & 19.87 & 5.78 \\
ACE++ & 1.174 & 0.570 & 87.09 & 29.92 & 13.25 \\
FireRed-Edit & 0.881 & 0.341 & 89.53 & 20.28 & 7.10 \\
Emu3.5 & 1.279 & 0.638 & 78.03 & 31.67 & 12.36 \\
ICEdit & 1.414 & 0.730 & 85.22 & 26.26 & 9.99 \\
Kling-v2.1-Autoreg & 1.020 & 0.303 & 30.30 & 20.93 & 6.55 \\
Seedance-1.0-Pro-Fast-Autoreg & 1.167 & 0.590 & 45.42 & 28.62 & 8.35 \\
\textit{HY-WorldMirror-2.0} & 1.491 & 0.512 & 10.51 & 2.52 & 1.35 \\
\textit{Lingbot-World} & 0.635 & 0.236 & 17.64 & 14.03 & 5.48 \\
\bottomrule
\end{tabular}
\caption{Full-pose diagnostics from the corrected DA3 records; lower is better. Eligible steps are averaged within case, then protocol, followed by an equal Atomic/Chain/Cycle macro. Translation direction and both rotation columns cover all eligible records. Full-vector translation fields cover 1{,}007 translation-action steps per model except GPT-Image-2, Seedream-5, and Gemini-3-Pro (887 each); missing vectors are left undefined rather than imputed. Italicized rows are pose-conditioned references.}
\label{tab:full_pose}
\end{table*}

These diagnostics explain why dominant-axis success is not equivalent to full-pose fidelity. HY-WorldMirror-2.0 has the smallest translation-direction and SO(3) errors, yet its full translation error is large because it often over-scales motion; conversely, a near-static model can have modest off-axis error while failing the instructed component. Among pose-free systems, GPT-Image-2 combines the strongest CMG with relatively low full-translation and rotation errors, but its mean translation-direction error remains $43.68^\circ$. The table is therefore diagnostic only: it does not enter CMG, SSP, Overall, or model ranking, and unequal full-vector coverage precludes treating small numerical differences as a separate leaderboard.

\subsection{Qualitative case studies}
\label{app:qualitative}

\begin{figure}[!htbp]
\centering
\includegraphics[width=0.84\textwidth]{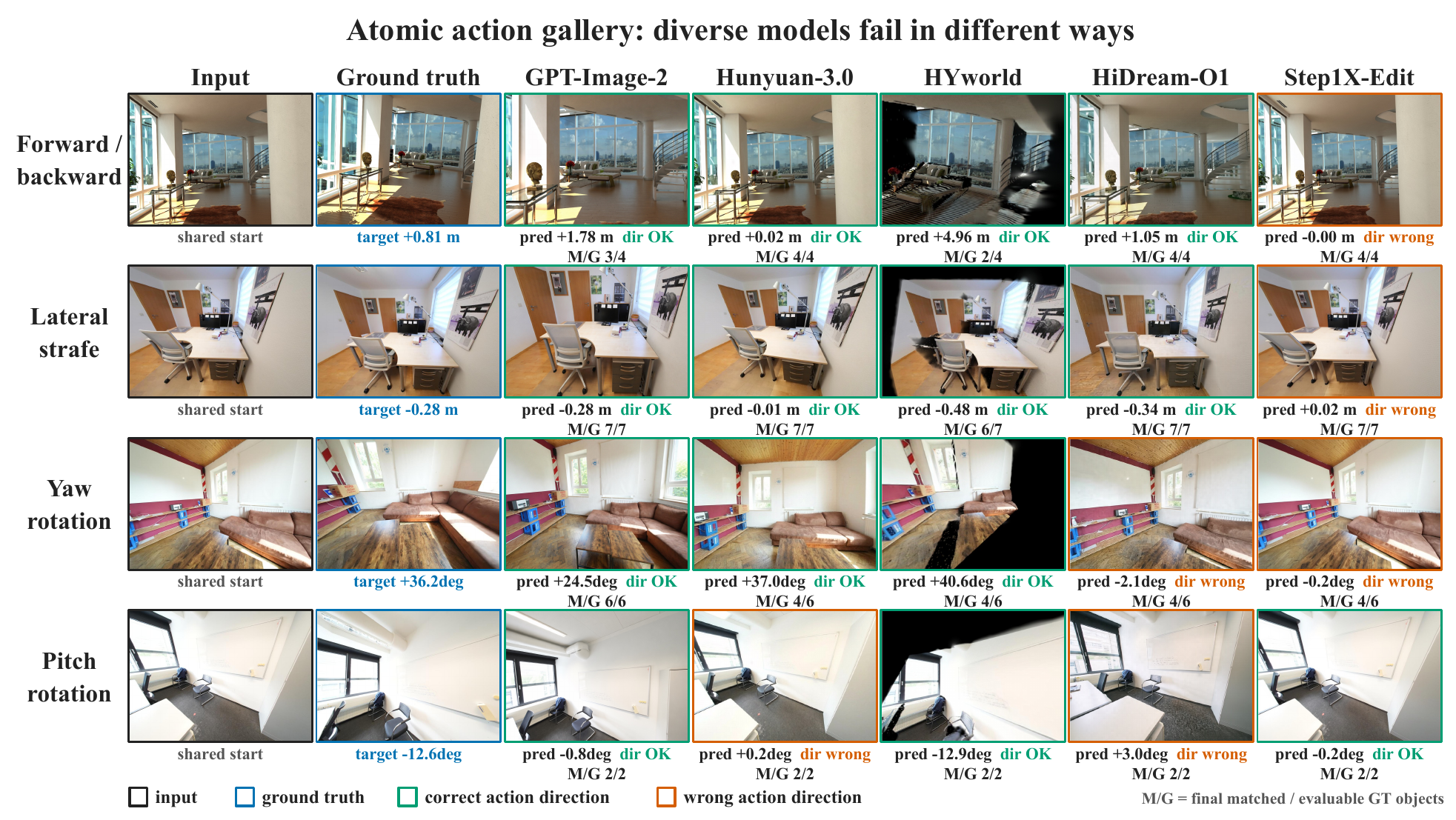}
\caption{Atomic action gallery. Rows are the four action families and columns are the shared input, the geometry-defined target, and five representative systems. Captions give the estimated dominant action, whether its direction matches the instruction, and the matched/evaluable GT-object count. Diverse systems fail in different ways: correct-direction magnitude errors, wrong-direction outputs, and object loss under degraded renders.}
\label{fig:atomic_gallery}
\end{figure}

\begin{figure}[!htbp]
\centering
\includegraphics[width=0.76\textwidth]{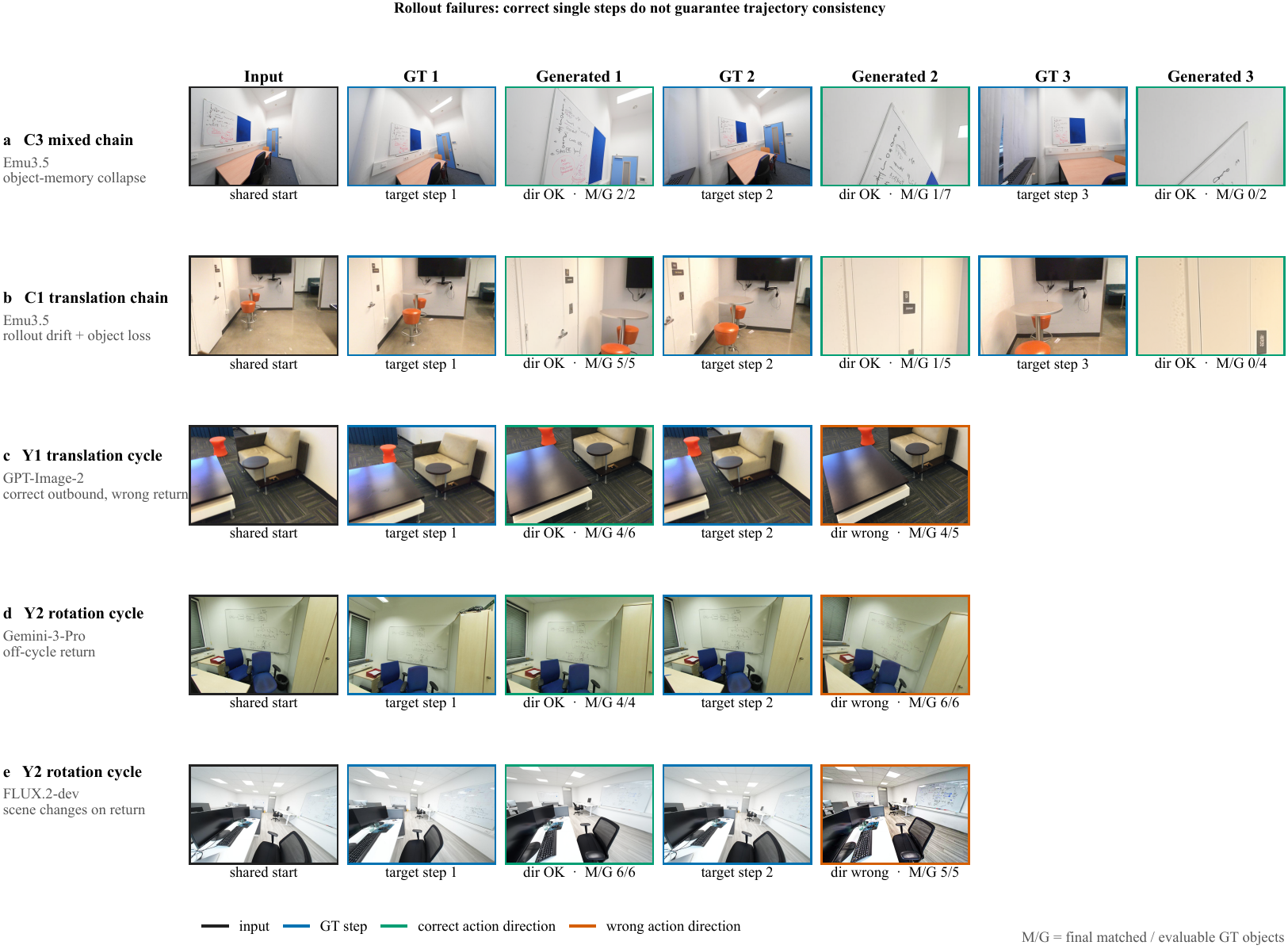}
\caption{Rollout gallery. Rows alternate the ground-truth and generated frames of three-step chains and two-step cycles. Chains expose cumulative object-retention loss as the matched/evaluable count drops across steps; cycles show cases in which a correct outbound motion is followed by an incorrect return. Correct single steps therefore do not guarantee trajectory consistency.}
\label{fig:rollout_gallery}
\end{figure}

Figures~\ref{fig:atomic_gallery} and~\ref{fig:rollout_gallery} illustrate the score behavior discussed above on concrete cases from the frozen bundle. Every panel reuses the official records: green and vermillion borders mark correct and wrong action direction, and each caption reports the estimated action and the matched-over-evaluable GT-object count (M/G) used by CMG and SSP.

Figure~\ref{fig:atomic_gallery} shows one atomic case per action family evaluated across five systems. A recurring pattern from the magnitude diagnostics in Table~\ref{tab:magnitude_compact} is visible directly: models frequently commit to the correct sense yet miss the requested magnitude, for example overshooting a forward translation or under-rotating a yaw, so the green border coexists with a large action error. Wrong-direction outputs and degraded renders lose object matches even when the scene is otherwise plausible.

Figure~\ref{fig:rollout_gallery} shows chain and cycle rollouts. The chain rows make the cumulative degradation of Section~\ref{app:diagnostics} concrete: individual steps can keep the correct direction while the matched-object count falls toward zero by the final step, so a trajectory drifts even when no single step is grossly wrong. The cycle rows show that a correct outbound step can be followed by a wrong-direction or weak return, consistent with the return-action failures in Appendix~\ref{app:cycle_behavior}.

\FloatBarrier

\section{\TrainSet{} as a Post-Training Resource}
\label{app:sft}

These experiments ask whether the benchmark can guide model development; they are separate from both the 16-system primary ranking and the pose-conditioned references. Every row is evaluated on all 1{,}400 cases/2{,}360 steps with the same CMG implementation, SSP exclusion mask, detector/matcher records, depth protocol, and aggregation as the main study. Each system produces one frozen output per step, and each SFT variant is trained once; the rows are therefore not averages over repeated training runs.

\subsection{Data construction and pair formation}

\TrainSet{} is mined from training-side scenes in DL3DV~\citep{ling2024dl3dv}, HyperSim, Matterport3D, ScanNet, and ScanNet++. The eligible source pool contains 8{,}315 scenes; geometry mining constructs the same four Atomic, three Chain, and two Cycle subtypes used by the benchmark. Candidate trajectories must satisfy the motion-dominance, overlap, visibility, depth-layer, and pose-separation gates. A subsequent visual and semantic filter rejects ambiguous motion, unusable images, different-scene sequences, dominant dynamic content, and cases without trackable spatial anchors. A final bidirectional depth-consistent check requires visible overlap in $[0.35,0.85]$ at every trajectory step. Exact $(\text{dataset},\text{scene ID})$ matching against the terminal blocklist leaves zero evaluation-scene overlap.

The final clean pool contains 4{,}140 scenes and 66{,}214 trajectories. Table~\ref{tab:sft_data_composition} separates trajectories from optimizer examples because each multi-step trajectory produces more than one edit pair.

\begin{center}
\begin{minipage}{\textwidth}
\centering
\captionsetup{hypcap=false}
\small
\setlength{\tabcolsep}{7pt}
\begin{tabular}{lrrr}
\toprule
\textbf{Protocol} & \textbf{Trajectories} & \textbf{Pairs per trajectory} & \textbf{Edit pairs} \\
\midrule
Atomic & 33{,}144 & 1 & 33{,}144 \\
Chain & 8{,}929 & 3 & 26{,}787 \\
Cycle & 24{,}141 & 2 & 48{,}282 \\
\midrule
Total & 66{,}214 & -- & 108{,}213 \\
\bottomrule
\end{tabular}
\captionof{table}{\TrainSet{} composition. Multi-step trajectories are converted into teacher-forced image-edit pairs while remaining intact during trajectory-level subset sampling.}
\label{tab:sft_data_composition}
\end{minipage}
\end{center}

The trajectory counts by source are 33{,}972 for DL3DV, 405 for HyperSim, 96 for Matterport3D, 6{,}627 for ScanNet, and 25{,}114 for ScanNet++. Instructions are generated from the calibrated relative pose and state the dominant action, direction, and approximate magnitude. At step $i>1$, teacher forcing uses the physical target from step $i-1$ as the first edit reference; it never uses a model prediction. Fixed same-scene auxiliary references remain available at every step, and current and target frames are excluded from auxiliary sampling.

The scale subsets are sampled from the final clean pool at the trajectory level with subtype stratification. The clean/raw comparison instead pairs a 40k sample from this final pool with a subtype-matched, best-effort sample from the raw geometry-mined pool before the visual/semantic and depth-consistent visibility filters. The raw pool contains too few C3 mixed chains to fill its requested quota, so the nominal raw-40k variant contains 39{,}761 trajectories and 64{,}653 edit pairs. This comparison tests the complete filtering pipeline; it does not isolate any single filtering rule.

\subsection{Qwen development probes}

Qwen-Image-Edit-2511 is fine-tuned after removing every evaluation scene. The full clean pool contains 66{,}214 trajectories and 108{,}213 teacher-forced edit pairs. Training uses eight NVIDIA A800 80-GB GPUs, global batch eight, BF16, gradient checkpointing, a 589{,}824-pixel dynamic-resolution cap, and rank-16 LoRA (\(\alpha=16\), zero dropout) on the diffusion transformer. AdamW uses learning rate \(10^{-4}\), weight decay 0.01, default betas, and frozen non-adapter weights.

Fixed-budget variants run for 4{,}000 updates. The full-clean continuation starts from the 4k LoRA weights and performs 9{,}527 additional updates, for 13{,}527 cumulative weight updates---approximately one pass over the 108{,}213 edit pairs. Because this continuation changes optimization budget, it is a development checkpoint rather than a controlled data-scaling result.

\begin{center}
\begin{minipage}{\textwidth}
\centering
\captionsetup{hypcap=false}
\small
\setlength{\tabcolsep}{3pt}
\begin{tabular}{lrrrrrr}
\toprule
\textbf{Variant} & \textbf{Train samples} & \textbf{Edit pairs} & \textbf{Budget} & \textbf{Overall} & \textbf{CMG} & \textbf{SSP} \\
\midrule
Base & -- & -- & no SFT & 0.495 & 0.470 & 0.520 \\
Clean 20k & 20{,}000 & 32{,}684 & 4k & 0.590 & 0.651 & 0.530 \\
Clean 40k & 40{,}000 & 65{,}370 & 4k & 0.589 & 0.664 & 0.513 \\
Clean full & 66{,}214 & 108{,}213 & 4k & 0.597 & 0.659 & 0.535 \\
Clean full continuation & 66{,}214 & 108{,}213 & $\sim$13.5k & 0.681 & 0.773 & 0.589 \\
Matched clean 40k & 40{,}000 & 65{,}370 & 4k & 0.615 & 0.698 & 0.531 \\
Matched raw 40k & 39{,}761 & 64{,}653 & 4k & 0.591 & 0.653 & 0.528 \\
\bottomrule
\end{tabular}
\captionof{table}{Benchmark-guided fine-tuning probe. Data-pool rows hold optimizer updates fixed; the continuation changes the training budget and is not a pure data-scale comparison.}
\label{tab:sft}
\end{minipage}
\end{center}

At 4k updates, increasing the cleaned pool from 20k to the full set does not produce a monotonic SSP trend, so these rows do not establish a scaling law. The per-protocol records show that Chain SSP rises from 0.397 for the base model to 0.442 for Clean full at 4k updates and to 0.505 after the 13.5k-update continuation. The single-step case improves far less: Atomic SSP rises only from 0.618 for the base model to 0.669 after the full continuation, even though this case gives the model a ground-truth previous frame and admits no rollout accumulation, and it does not rise as the pool grows. The full continuation reaches 0.681 Overall but uses substantially more optimization. These point estimates establish that benchmark-derived data can produce a competitive held-out-scene checkpoint; the matched controls below are required to attribute axis-specific differences.

\subsection{Matched endpoint and second-backbone controls}
\label{app:sft_matched}

\paragraph{Training and generation controls.}
The Qwen comparison continues the independently trained clean/raw 40k adapters for exactly 9{,}527 additional local updates, yielding 13{,}527 updates from the same base for both endpoints. Continuation restores adapter weights only and reinitializes optimizer and scheduler state; the original 4k adapters were trained without an explicit seed. The comparison is therefore update-matched, not bitwise-equivalent to uninterrupted training and not averaged over training randomness.

For the second backbone, we convert the same 65{,}370 cleaned 40k Qwen edit pairs one-to-one to the OmniGen2 format without resampling, reordering, or prompt rewriting. OmniGen2 trains for 4{,}000 updates on eight A800 80-GB GPUs with global batch eight, BF16, gradient checkpointing, seed 2233, and rank-8 LoRA (\(\alpha=8\), zero dropout) on attention query/key/value/output projections. AdamW uses learning rate \(8\!\times\!10^{-7}\), betas \((0.9,0.95)\), weight decay 0.01, gradient clipping at one, and a 500-update warmup. The matched base and SFT generations use the same benchmark inputs, seed 42, sampler, target-size rule, and strict Cycle step-2 conditioning. Qwen generation uses $1024^2$ outputs; OmniGen2 uses 50 Euler steps, text guidance 5.0, and image guidance 2.0.

All four matched rows contain 1{,}400 cases and 2{,}360 steps with zero missing outputs and identical validity masks. Reconstructed headline and pillar scores match the frozen final metric files to below \(7\!\times\!10^{-16}\) absolute error.

\begin{center}
\begin{minipage}{\textwidth}
\centering
\captionsetup{hypcap=false}
\small
\setlength{\tabcolsep}{7pt}
\begin{tabular}{lrrr}
\toprule
\textbf{Matched endpoint} & \textbf{Overall} & \textbf{CMG} & \textbf{SSP} \\
\midrule
Qwen raw 40k, 13.5k updates & 0.669 & 0.766 & 0.572 \\
Qwen clean 40k, 13.5k updates & 0.673 & 0.783 & 0.563 \\
OmniGen2 matched base & 0.449 & 0.440 & 0.457 \\
OmniGen2 clean-40k SFT, 4k updates & 0.430 & 0.450 & 0.410 \\
\bottomrule
\end{tabular}
\captionof{table}{Absolute scores for the matched SFT controls. The Qwen rows isolate clean versus raw training data at the same cumulative update count; the OmniGen2 base is regenerated under the SFT row's inference configuration.}
\label{tab:sft_matched_absolute}
\end{minipage}
\end{center}

\paragraph{Paired uncertainty.}
The primary analysis uses 20{,}000 paired scene-cluster bootstrap resamples (seed 20260810): scenes are sampled with replacement within each source dataset, all cases inherit their scene multiplicity, and the full protocol macro is recomputed before subtraction. A 20{,}000-resample subtype-stratified case bootstrap gives the same headline decisions. Intervals characterize paired benchmark sampling for these frozen runs, not decoding or training-run variance.

\begin{center}
\begin{minipage}{\textwidth}
\centering
\captionsetup{hypcap=false}
\scriptsize
\setlength{\tabcolsep}{3.5pt}
\begin{tabular}{lrrrr}
\toprule
\textbf{Paired comparison} & $\boldsymbol{\Delta}$\textbf{Overall} & $\boldsymbol{\Delta}$\textbf{CMG} & $\boldsymbol{\Delta}$\textbf{SSP} & $\boldsymbol{\Delta}$\textbf{CMG}$-\boldsymbol{\Delta}$\textbf{SSP} \\
\midrule
Qwen clean $-$ raw, 13.5k & $+0.004$ [$-0.003,+0.011$] & $+0.017$ [$+0.007,+0.026$] & $-0.009$ [$-0.019,+0.002$] & $+0.025$ [$+0.011,+0.040$] \\
OmniGen2 SFT $-$ base & $-0.019$ [$-0.031,-0.008$] & $+0.010$ [$-0.010,+0.029$] & $-0.048$ [$-0.059,-0.037$] & $+0.057$ [$+0.036,+0.080$] \\
\bottomrule
\end{tabular}
\captionof{table}{Paired differences with 95\% scene-cluster bootstrap intervals. Qwen cleaning reliably improves CMG but not Overall at the long endpoint; OmniGen2 SFT leaves CMG unresolved while reliably reducing SSP and Overall. The positive axis-gap intervals are the cross-backbone evidence that pairwise supervision changes action and state at different rates.}
\label{tab:sft_paired_effects}
\end{minipage}
\end{center}

For Qwen, both CMG components support the action gain: direction changes by $+0.019$ [$+0.008,+0.030$] and gated magnitude by $+0.015$ [$+0.005,+0.024$]. SSP F1 decreases by $-0.012$ [$-0.023,-0.001$], whereas spatial and integrity changes remain unresolved. For OmniGen2, SSP F1, spatial relations, and integrity fall by $-0.037$, $-0.058$, and $-0.049$, respectively, with all intervals below zero. Hence the second backbone supports an action--state divergence, not a claim that SFT improves CMG.

\begin{center}
\begin{minipage}{\textwidth}
\centering
\captionsetup{hypcap=false}
\scriptsize
\setlength{\tabcolsep}{4pt}
\begin{tabular}{llrr}
\toprule
\textbf{Comparison} & \textbf{Protocol} & $\boldsymbol{\Delta}$\textbf{CMG} & $\boldsymbol{\Delta}$\textbf{SSP} \\
\midrule
\multirow{3}{*}{Qwen clean $-$ raw} & Atomic & $-0.001$ [$-0.006,+0.004$] & $+0.003$ [$-0.011,+0.016$] \\
 & Chain & $+0.036$ [$+0.017,+0.056$] & $-0.011$ [$-0.028,+0.005$] \\
 & Cycle & $+0.015$ [$-0.003,+0.033$] & $-0.017$ [$-0.040,+0.006$] \\
\midrule
\multirow{3}{*}{OmniGen2 SFT $-$ base} & Atomic & $+0.004$ [$-0.023,+0.031$] & $-0.026$ [$-0.040,-0.011$] \\
 & Chain & $+0.030$ [$+0.001,+0.058$] & $-0.056$ [$-0.073,-0.040$] \\
 & Cycle & $-0.005$ [$-0.046,+0.036$] & $-0.062$ [$-0.088,-0.036$] \\
\bottomrule
\end{tabular}
\captionof{table}{Protocol-level paired differences and 95\% scene-cluster intervals. Qwen's reliable CMG gain localizes to Chain, while OmniGen2 SSP declines under every protocol and most strongly in sequential settings. Protocol rows are diagnostic decompositions of the headline effects.}
\label{tab:sft_protocol_effects}
\end{minipage}
\end{center}

\subsection{Budget, subgroup, and qualitative diagnostics}

\paragraph{Budget.}
The early Qwen cleaning advantage does not grow under continued optimization. At 4k updates, clean minus raw is $+0.024$ [$+0.015,+0.033$] Overall and $+0.045$ [$+0.032,+0.058$] CMG, but this advantage contracts at the long endpoint until Overall is no longer separated from zero. SSP is $+0.003$ [$-0.008,+0.014$] at 4k and $-0.009$ [$-0.019,+0.002$] at 13.5k. Combined with the fixed-update 20k/40k/full-pool probe, this rules out the simple prescription that exposing the same pairwise objective to more distinct pairs or more updates is sufficient for state preservation.

\paragraph{Source, subtype, action, and concentration.}
Source-level estimates are heterogeneous and source is not fully crossed with protocol. Qwen's resolved SSP loss appears in ScanNet++ ($-0.012$ [$-0.023,-0.002$]); OmniGen2 SSP falls on ScanNet ($-0.046$ [$-0.075,-0.017$]) and ScanNet++ ($-0.043$ [$-0.055,-0.031$]), while the much smaller Matterport3D slice is inconclusive. The subtype map shows Qwen's largest axis gap in C1 translation chains and OmniGen2 gaps across translation, rotation, and mixed sequential subtypes. In an explicitly exploratory step-level action slice, forward motion gives Qwen $+0.072$ CMG and $+0.013$ SSP, but OmniGen2 $+0.083$ CMG and $-0.052$ SSP. The top 10\% of scenes account for at most 43.9\% of absolute headline contribution across the four model--metric combinations, so no headline effect is carried by a handful of scenes. Dataset, subtype, and action intervals are uncorrected diagnostics rather than independent discoveries.

\paragraph{Prespecified qualitative audit.}
Before visual inspection, we select two distinct-scene cases for each comparison in each of three categories: joint improvement ($\Delta$CMG and $\Delta$SSP both above $0.05$), action--state tradeoff ($\Delta$CMG above $0.05$ and $\Delta$SSP below $-0.05$), and joint regression (both below $-0.05$). Cases are ranked deterministically within category, and every input, GT target, and generated step is retained. The full audit contains 12 cases. In the top tradeoff case for each backbone (Qwen Cycle, $\Delta$CMG $=+0.502$ and $\Delta$SSP $=-0.323$; OmniGen2 Atomic, $\Delta$CMG $=+0.956$ and $\Delta$SSP $=-0.491$), the treatment executes the requested motion more successfully but preserves fewer target-view objects or relations, so a positive Overall change ($+0.089$ for Qwen; $+0.232$ for OmniGen2) hides a large SSP loss. These metric-extreme examples illustrate the exchange and are not frequency estimates.

\paragraph{Interpretation boundary.}
The completed controls support the conclusion that pairwise SFT changes the two axes at different rates. They do not support universal data-cleaning gains, an OmniGen2 CMG improvement, or a causal claim averaged over training seeds. Although the data contain Chain and Cycle trajectories, teacher forcing conditions every later step on the preceding physical target and therefore does not train recovery from model-induced drift. A stronger baseline should select checkpoints on the CMG--SSP Pareto frontier or enforce SSP non-degradation, then test explicit identity/count, planar/depth topology, integrity, free-running sequence, and worst-step losses under matched compute.

\FloatBarrier

\section{Reference Baselines and Score Calibration}
\label{app:references}

Four unranked references calibrate the metric scale without defining mathematical lower or upper bounds. GT-target oracle emits the physical target at every step. Source-copy recursively emits the initial view without executing an action. Oracle nearest-visible-input copy uses the GT target only to select the most similar image among the current and auxiliary views exposed by the benchmark interface; it never inserts the target into the candidate set. Scene-disjoint NN retrieval-copy selects a training transition after removing all 771 evaluation scenes, then restricts candidates by $K$, action direction, and a training-only magnitude tertile before CLIP-based selection. No evaluation target image, feature, pose, depth, mask, or label enters the scene-disjoint retrieval procedure.

\begin{center}
\begin{minipage}{\textwidth}
\centering
\captionsetup{hypcap=false}
\small
\setlength{\tabcolsep}{4pt}
\begin{tabular}{lrrr}
\toprule
\textbf{Reference} & \textbf{Overall} & \textbf{CMG} & \textbf{SSP} \\
\midrule
GT-target oracle & 0.940 & 0.980 & 0.899 \\
Source-copy & 0.391 & 0.181 & 0.601 \\
Oracle nearest-visible-input copy & 0.457 & 0.302 & 0.613 \\
Scene-disjoint NN retrieval-copy & 0.377 & 0.539 & 0.215 \\
\bottomrule
\end{tabular}
\captionof{table}{Unranked score-calibration references. CMG uses the model-independent 2{,}358-step reference mask; SSP uses the same 44-case exclusion mask as the primary evaluation.}
\label{tab:reference_baselines}
\end{minipage}
\end{center}

The ranked set contains 16 pose-free systems; HY-WorldMirror-2.0 and Lingbot-World remain separate pose-conditioned references. GPT-Image-2 reaches 0.662 Overall, leaving a 0.278 gap to the practical oracle. Source-copy shows that preserving the input without moving can retain a moderate SSP but receives little CMG credit. The two retrieval rows are diagnostics rather than deployable methods: oracle nearest-visible-input copy uses the target only to select among already exposed views, whereas scene-disjoint NN retrieval-copy never accesses evaluation targets or scenes.

\section{Reproduction Scope}
\label{app:reproduction}

The planned release is designed around stable case/step identifiers that connect permitted model inputs and instructions with evaluator-only targets, poses, depth, visibility, and object metadata. Pose-free inference reads only the permitted images and language instruction. Versioned manifests will record file hashes and the fixed SSP mask; accompanying scripts will recompute CMG, SSP, aggregation, human-alignment statistics, bootstrap intervals, and claim-focused diagnostics from frozen outputs and per-step records. The evaluation release will cover the selected 1{,}400 cases; exhaustive intermediate candidates from early mining are not required to reproduce the reported evaluation. For \TrainSet{}, the release scope comprises the row-level trajectory manifest, source-relative frame identifiers, pose-derived instructions, deterministic teacher-forced conversion, construction metadata, frozen hashes, and the exact scene-leakage audit. Restricted source pixels, private absolute paths, and third-party model weights are not redistributed.

\section{Evaluator Configuration and Compute}
\label{app:compute}

Table~\ref{tab:evaluator_config} lists the frozen models that implement the two metrics and the auxiliary diagnostics. Every component runs with fixed weights and fixed hyperparameters, and no evaluator is trained or fine-tuned on benchmark outputs. Given the frozen generated images and per-step records, the pipeline is deterministic, so re-running it reproduces the reported scores exactly.

\begin{table}[t]
\centering
\small
\begin{tabularx}{\columnwidth}{@{}>{\raggedright\arraybackslash\hsize=0.72\hsize}X>{\raggedright\arraybackslash\hsize=1.28\hsize}X@{}}
\toprule
\textbf{Component} & \textbf{Role in evaluation} \\
\midrule
DA3Nested-Giant-Large~\citep{lin2025depth3recoveringvisual} & Relative camera pose for CMG; single-image depth for SSP depth topology \\
Grounding DINO~\citep{liu2024grounding} & Open-vocabulary object detection in GT and generated images \\
Qwen3-VL~\citep{bai2025qwen3vltechnicalreport} & GT target-label proposal and propose--demote--recover matching \\
DINOv3~\citep{simeoni2025dinov3} & Appearance-identity verification for matched objects \\
CLIP ViT-L/14@336 & Feature-space view-confusion diagnostic (Context-$K$) \\
VGGT-1B~\citep{wang2025vggt} & Alternate pose/depth backend for robustness ablations \\
\bottomrule
\end{tabularx}
\caption{Frozen evaluator components and their roles. All run with fixed weights; SAM3 segmentation is disabled, so SSP uses rectangular detector boxes.}
\label{tab:evaluator_config}
\end{table}

Evaluator inference runs on a single 80-GB GPU per component, and the detection, matching, and pose stages dominate wall-clock time. Generation cost is not separately metered here because it varies by provider and interface: closed systems (GPT-Image-2, Seedream-5, Gemini-3-Pro) are queried through their public APIs, whereas open systems run locally under their native decoding settings. We therefore report evaluator configuration rather than a single normalized cost figure. The fine-tuning probe in Section~\ref{app:sft} uses eight NVIDIA A800 80-GB GPUs, as stated there.

\section{Data Licensing and Ethics}
\label{app:ethics}

\Bench{} is assembled from four established research RGB-D datasets---ScanNet++~\citep{yeshwanth2023scannet++}, ScanNet~\citep{dai2017scannet}, HyperSim~\citep{roberts2021hypersim}, and Matterport3D~\citep{chang2017matterport3d}. \TrainSet{} additionally uses DL3DV~\citep{ling2024dl3dv} as a training-only source. Both resources are used for non-commercial academic research under the original licenses and access terms. No new scenes are captured for this work. Source scans may contain sensitive visual details inherited from the provider datasets. The release plan includes a reporting and removal route for affected records; we do not assume that the upstream datasets are free of person-specific content.

\paragraph{Release boundary.}
The planned release will provide versioned metadata, source-relative identifiers, scene-disjoint split records, and local materialization scripts rather than republishing restricted source pixels. Users will obtain the underlying datasets from their official providers and remain subject to the corresponding licenses and access agreements.

The human validation study in Section~\ref{app:human} uses rubric-trained annotators who rank anonymized system outputs and collect only ordinal quality judgments, with no personal data recorded. Released material links cases to source frames through stable identifiers and provides construction metadata and per-step evaluator records needed to reproduce the reported results; redistribution of the underlying scans follows each source dataset's own terms rather than republishing them. Because the benchmark targets physical camera-motion consistency, it is intended as a diagnostic for controllable and world-model generation and carries the standard dual-use considerations of generative-model research; it adds no generative capability beyond evaluation.

\section{Scope and Interpretation Boundaries}
\label{app:limitations}

\begingroup
\setlength{\parskip}{3pt}
\titlespacing*{\paragraph}{0pt}{3pt}{0.6em}

\paragraph{Scope.}
\Bench{} evaluates static indoor scenes, four dominant camera actions, and rollouts of at most three steps; it does not cover dynamics, physical interaction, long-horizon memory, or open-world navigation.

\paragraph{Evaluator dependence.}
CMG depends on learned relative-pose estimation and per-transition GT-assisted translation-scale calibration. Direction-only and backend tests support broad rank stability, but magnitude conclusions remain calibration-dependent. SSP depends on frozen detection, target-derived prompts, correspondence, and monocular depth. Its mean-plus-worst aggregation intentionally emphasizes intermediate failure; matched aggregation shows that the large official CMG/SSP Chain-drop contrast is not invariant to the temporal functional. Ablations support broad ranks, while close ranks and Cycle correspondence remain evaluator-sensitive.

\paragraph{Observable objects.}
SSP covers only objects supported by the supplied views and formal visibility/matchability filters; it is not an exhaustive open-world inventory, and an unmatched detection is not proof of physical absence.

\paragraph{Human evidence.}
Human validation covers a six-system panel rather than all 18 evaluated systems. Agreement is strongest after case/protocol aggregation, so the metrics support aggregate system comparison rather than exact per-example grading.

\paragraph{Attribution and context.}
SSP is a joint target-view outcome that can fall through scene corruption, wrong pose, visibility, or evaluator error; pose-conditioned strata are observational and do not isolate state memory. The fixed-object Context-$K$ decline does not causally estimate the value of extra context because interfaces, evidence sets, and behavior remain intertwined.

\paragraph{Evaluation scope and interfaces.}
Bootstrap intervals characterize benchmark sampling for one frozen output per model--step, not decoding variation. Source-removal analyses support broad rank stability, but source coverage is imbalanced and Matterport3D has only 22 cases. The benchmark compares complete systems through predefined native interfaces, whose image limits, resolutions, and decoding remain part of each system configuration.

\paragraph{Training probe.}
The SFT study contains one training run per variant. The Qwen clean/raw adapters were independently trained without an explicit seed, their continuations restore adapter weights without optimizer state, and the OmniGen2 matched base controls generation configuration rather than training randomness. Bootstrap intervals quantify paired benchmark sampling only. The two backbones support axis-selective score changes under pairwise supervision; repeated-seed claims and comparisons with free-running, state-aware, or full-pose objectives require matched experiments. Because later training steps use preceding physical targets, the present data do not establish how the same objectives behave under model-generated histories.

\paragraph{Data provenance and contamination.}
Evaluation scenes are disjoint from the SFT data; pretraining exposure to the source datasets remains unknown.

\paragraph{Reading model ranks.}
Scene-cluster resampling leaves several adjacent systems tied, so scores should be read as capability profiles and broad tiers, not exact ranks. Weight and sample-efficiency tests do not guarantee future rankings; pose-conditioned references receive privileged 6-DoF controls and remain separate.
\endgroup

\end{document}